\pdfoutput=1
\documentclass[10pt,journal]{IEEEtran}
\IEEEoverridecommandlockouts 
\usepackage{blkarray}                                      
\usepackage{algpseudocode}                                 
\usepackage{algorithm}
\usepackage{graphicx}                                      
\usepackage{amsmath}
\usepackage{amssymb}
\usepackage{amsfonts}
\usepackage{amsthm}
\usepackage[mathcal]{eucal}
\usepackage{mathrsfs}
\usepackage{booktabs}
\usepackage{enumerate}
\usepackage{multirow}
\usepackage[subrefformat=parens,farskip=0pt,justification=centering]{subfig}
\usepackage{color}
\usepackage{cite}                                          
\usepackage{comment}                                       
\usepackage{soul}                                          
\soulregister\cite7
\soulregister\ref7
\soulregister\pageref7
\usepackage{etoolbox}                                      
\usepackage{url}
\usepackage{nth}                                           
\usepackage{bm}                                            
\usepackage{courier}
\usepackage{balance}
\usepackage{threeparttable}
\usepackage{xcolor,colortbl}
\usepackage{footnote}
\usepackage{listings}
\usepackage{setspace}                                      
\usepackage[inline]{enumitem}

\usepackage{verbatim}
\usepackage[bookmarks=false]{hyperref}
\hypersetup{
    colorlinks = true,
    citecolor  = blue,
    linkcolor  = blue,
    urlcolor   = blue,
}
\usepackage{tikz}
\usetikzlibrary{patterns,snakes}
\usetikzlibrary{positioning,calc,fit,decorations.pathmorphing,shapes.geometric, shapes.gates.logic.US, calc}
\usetikzlibrary{arrows,arrows.meta,decorations.markings,shapes,shapes.arrows}
\usetikzlibrary{decorations,decorations.pathreplacing}
\usetikzlibrary{backgrounds}
\usepackage{filecontents}                                  
\usepackage{pgfplots}
\usepackage{pgfplotstable}
\usepackage{scalefnt}
\pgfplotsset{compat=newest}
\usepackage{caption}
\usepackage{cleveref}
\Crefformat{figure}{Fig.~#2#1#3}                           
\Crefname{subfigure}{Fig.}{Figs.}
\Crefname{figure}{Fig.}{Figs.}
\Crefformat{table}{TABLE~#2#1#3}                           
\usepackage[figuresright]{rotating}

\definecolor{CUHKorange}{RGB}{244,106,18} 
\definecolor{CUHKblue}{RGB}{0,111,190}    
\definecolor{CUHKgreen}{RGB}{0,127,128}   
\definecolor{CUHKred}{RGB}{228,46,36}     
\definecolor{CUHKyellow}{RGB}{198,148,34} 
\definecolor{CUHKdark}{RGB}{114,44,114}   
\definecolor{CUHKmiddle}{RGB}{144,44,144} 
\definecolor{CUHKlight}{RGB}{167,44,167} 
\definecolor{CUHKpurple}{RGB}{117,15,109}
\definecolor{CUHKgold}{RGB}{221,163,0}
\definecolor{CUHKribbon}{RGB}{244,223,176}
\definecolor{CUHKblack}{RGB}{34,24,21}

\renewcommand{\bf}[1]{\textbf{#1}}
\renewcommand{\tt}[1]{\texttt{#1}}
\renewcommand{\vec}[1]{\boldsymbol{#1}}    

\usepackage{tcolorbox}
\tcbuselibrary{skins,breakable}
    {\endtcolorbox}
    {\endtcolorbox}

\usepackage[top=0.78in,bottom=0.86in,left=0.62in,right=0.62in]{geometry}
\newtheorem{myproblem}{\textbf{Problem}}
\newtheorem{mydefinition}{\textbf{Definition}}

\crefname{mytheorem}{Theorem}{Theorems}
\crefname{mylemma}{Lemma}{Lemmas}
\crefname{myclaim}{Claim}{Claims}
\crefname{myproperty}{Property}{Properties}
\crefname{mycorollary}{Corollary}{Corollaries}

\algrenewcommand\textproc{\texttt}

\makeatletter
\let\OldStatex\Statex
\renewcommand{\Statex}[1][3]{%
  \setlength\@tempdima{\algorithmicindent}%
  \OldStatex\hskip\dimexpr#1\@tempdima\relax
}
\makeatother

\RequirePackage[normalem]{ulem} 
\RequirePackage{color}\definecolor{RED}{rgb}{1,0,0}\definecolor{BLUE}{rgb}{0,0,1} 

\graphicspath{{./figs/}{../}}
\definecolor{myorange}{RGB}{238,97,42}  %
\definecolor{myblue}{RGB}{178,179,249}  
\definecolor{mygrey}{RGB}{166,166,166}  %
\definecolor{mygreen}{RGB}{180,210,36}  
\definecolor{myred}{RGB}{238,0,0}       
\definecolor{myyellow}{RGB}{198,148,34} 
\definecolor{mydark}{RGB}{114,44,114}   
\definecolor{mymiddle}{RGB}{144,44,144} 
\definecolor{mylight}{RGB}{167,44,167}  
\definecolor{myblue1}{RGB}{137,157,192}  
\definecolor{mygreen1}{RGB}{69,137,148}  

\definecolor{CUpurple}{RGB}{136,43,142}
\definecolor{CUlpurple}{RGB}{165,133,182}
\definecolor{CUgold}{RGB}{221,163,0}
\definecolor{CUribbon}{RGB}{244,223,176}

\definecolor{myblue1}{RGB}{137,157,192}  
\definecolor{mygreen1}{RGB}{69,137,148}  
\definecolor{mygreen2}{RGB}{0,174,75} 
\definecolor{myred1}{RGB}{254,189,189} 
\definecolor{xzblue}{RGB}{188,211,224}
\begin{document}


\title{
	Learning-driven Physically-aware Large-scale Circuit Gate Sizing 
}

\author
{
    Yuyang Ye,     \quad
    Peng Xu,       \quad
    Lizheng Ren,   \quad
    Tinghuan Chen, \quad
    Hao Yan,       \quad
    Bei Yu,        \quad
    Longxing Shi\\

    \thanks{This work was submitted to IEEE Transactions on Computer-Aided Design of Integrated Circuits and Systems (\textbf{TCAD}).}
    \thanks{Yuyang Ye, Hao Yan, Lizheng Ren, and Longxing Shi are with the National ASIC Research Center, Southeast University, Nanjing 210096, China, also with the National Center of Technology Innovation for EDA, Nanjing, China.}
    \thanks{Peng Xu and Bei Yu are with the Department of Computer Science and Engineering, The Chinese University of Hong Kong, NT, Hong Kong SAR.}
    \thanks{Tinghuan Chen is with the School of Science and Engineering, 
The Chinese University of Hong Kong, Shenzhen, China.}
}

\maketitle
\pagestyle{plain}

\begin{abstract}
Gate sizing plays an important role in timing optimization after physical design.
Existing machine learning-based gate sizing works cannot optimize timing on multiple timing paths simultaneously and neglect the physical constraint on layouts.
They cause sub-optimal sizing solutions and low-efficiency issues when compared with commercial gate sizing tools.
In this work, we propose a learning-driven physically-aware gate sizing framework to optimize timing performance on large-scale circuits efficiently.
In our gradient descent optimization-based work, for obtaining accurate gradients, a multi-modal gate sizing-aware timing model is achieved via learning timing information on multiple timing paths and physical information on multiple-scaled layouts jointly.
Then, gradient generation based on the sizing-oriented estimator and adaptive back-propagation are developed to update gate sizes.
Our results demonstrate that our work achieves higher timing performance improvements in a faster way compared with the commercial gate sizing tool.
\end{abstract}
\section{Introduction}
\begin{figure}[!t]
    \subfloat[Classical gate sizing flow where many iterations are necessities.]{\centering\includegraphics[width=0.99\linewidth]{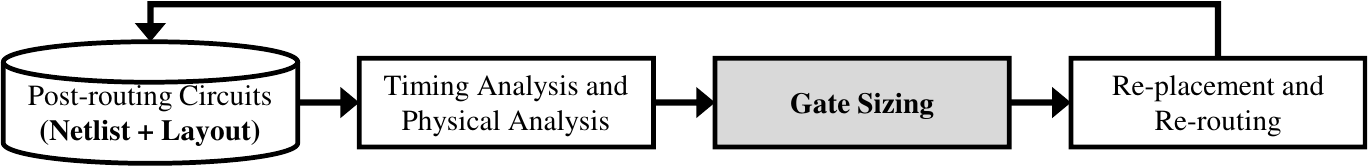} \label{fig:flow} } \\
    \subfloat[Netlist with multiple paths.]{\centering\includegraphics[width=0.55\linewidth]{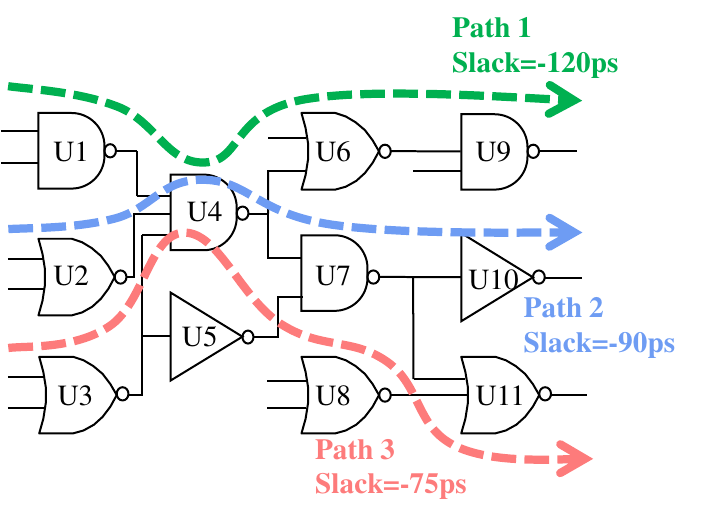} \label{fig:netlist} }
    \subfloat[Layout under multiple scales.]{\centering\includegraphics[width=0.4\linewidth]{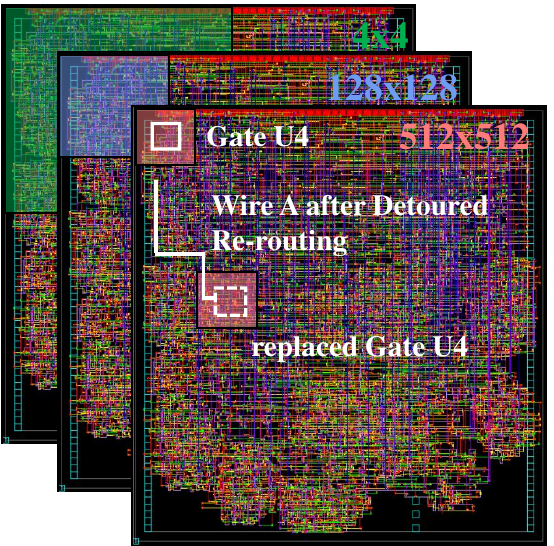} \label{fig:layout} }
    \caption{Rich information (a) in optimization flow; (b) on timing paths; (c) on design layouts.  } 
    \label{fig:path-node}
\end{figure}

Gate sizing on post-routing circuits is fundamental for timing optimization to achieve sign-off timing closure with smaller the worst negative slack (WNS) and total negative slack (TNS).
The solution space scales exponentially with respect to the size of circuits \cite{kahng2011vlsi,kahng2013high}.
Under advanced technology, as illustrated in \Cref{fig:flow}, physically-aware timing ECO flow is proposed.
The flow can consider physical information and timing information jointly to achieve timing closure \cite{icc2}.
However, poor convergence forces engineers to perform many time-consuming iterations throughout the flow \cite{nath2022transsizer}. 
It makes an efficiency bottleneck for gate sizing.

Existing gate sizing algorithms can be divided into two kinds.
(1) Analytical methods \cite{kahng2013high,livramento2014hybrid,sharma2015fast,sharma2019fast,roy2015osfa,mangiras2022task,daboul2018provably}:
discrete gate sizing is solved through gradient descent optimization using Lagrangian relaxation-based algorithms in these methods. 
(2) Machine-learning methods \cite{lu2021rl,nath2022transsizer,cheng2023dagsizer}:
machine learning models are used to perform gate sizing through modeling circuits.
Although machine learning has achieved many improvements in the gate-sizing problem, the performance of previous works cannot meet industry requirements when compared with commercial EDA tools.
In RL-sizer \cite{lu2021rl}, the generalization ability and runtime costs limit the application.
In Transizer \cite{nath2022transsizer}, optimization performance is sensitive to the accuracy of the proposed gate sizing prediction model.
A small prediction error that happens on critical paths always causes terrible optimization results, which makes Transizer difficult to achieve stable and really optimal performance.

Recently, learning-driven gradient descent optimization works have solved some EDA issues \cite{liu2021global,liu2023concurrent,pham2023agd,chen2024ultra,zhu2023l2o,guo2022differentiable}.
Fortunately, it is also a good idea for gate-sizing which combines analytical and machine learning methods jointly.
However, totally different from other EDA problems, it is a special task to achieve gate-sizing based on learning-driven gradient descent optimization.
There are two main challenges:
(1) achieving a gate sizing-aware timing model where accurate optimization gradients are calculated based on it.
(2) generating and back-propagating gradients w.r.t. discrete gate sizes on large-scale circuits efficiently and effectively.

For challenge (1), modeling gate sizing-induced timing performance variations urgently needs timing information on paths and physical information on layouts.
For timing paths, the path delay variations of multiple timing paths caused by gate sizing on one single gate always are different \cite{kahng2013high}.
As shown in \Cref{fig:netlist}, Path 1 and Path 2 are critical paths that go through gate $\operatorname{U4}$.
The setup timing performance of Path 1 can be optimized through upsizing gate $\operatorname{U4}$.
However, the larger effective capacitance of up-sized $\operatorname{U4}$ loaded on gate $\operatorname{U2}$ and $\operatorname{U1}$ causes delay degradation on Path 2.
The trade-off between the optimization and degradation on multiple critical paths should be achieved while gate sizing.
On design layouts, gate sizing might cause wire delay degradations after re-placement and re-routing.
As shown in \Cref{fig:layout}, when replacement happens on gate $\operatorname{U4}$ in the region with high gate density, up-sized gate $\operatorname{U4}$ must be replaced to avoid overlapping on layouts.
In the re-routing stage, on the layout with high wire congestion, the wire length of wire A increases due to detoured routing.
After that, the delay of wire A degrades with the wire lengths.
For layouts under multiple scales, the results of re-placement and re-routing are different \cite{kahng2011vlsi}.
Thus, physical information on multiple-scaled layouts is important to trade off wire delay degradations.
In addition, different from prediction works, the target of the timing model used in our work is to achieve optimal gate-sizing.
The optimization information from the commercial gate sizing tool can be considered to guide the gradient.
In summary, timing information on paths, physical information on layouts and optimization information in the industrial flow should be given full and joint consideration.

For challenge (2), the size of each gate is discrete rather than continuous.
It means $\operatorname{round}$ functions should be used in timing models based on achievable gate sizes.
However, $\operatorname{round}$ functions are not differentiable.
AGD \cite{pham2023agd} proposed to use the $\operatorname{Softmax}$ functions to replace $\operatorname{round}$ functions for approximating the gradients in discrete functions as categorical variables.
However, the gate size should be regarded as an integer-valued variable to retain the relationships between different sizes.
Inspired by recent quantization-aware training works, the straight-through estimator can help us to obtain accurate gradients w.r.t. discrete gate sizes \cite{yin2019understanding,le2022adaste,yang2022injecting}.
It is helpful to avoid discrepancies between the forward and backward pass, leading to global optimal gate sizing results.
In addition, on large-scale circuits with numerous gates, there is a high-dimensional issue during gradient back-propagation.
When numerous gates update sizes simultaneously, there are interdependencies among them.
Limited considerations about the problem cause low optimization efficacy.

In this work, we propose a learning-driven physically-aware gate sizing framework to achieve timing optimization on large-scale circuits efficiently.
Our work overcomes the above challenges of achieving gate sizing via gradient descent optimization.
To obtain a gate sizing-aware timing model, we learn optimization information, timing information on paths and physical information on layouts jointly through multi-modal learning.
The learned information helps to accurately model timing optimization and degradation induced by gate sizing.
To update gate sizes based on gradients effectively, we generate accurate timing performance gradients w.r.t. integer-valued gate sizes and back-propagate them with different priorities on different gates.
We highlight our contributions:
\begin{itemize}
    \item For the first time, we propose a learning-driven framework to achieve physically-aware gate-sizing.
    It can optimize timing performance effectively on large-scale circuits.
    \item We achieve multi-modal gate sizing-aware timing modeling via timing information aggregation on multiple critical paths and physical information aggregation on multiple scaled layouts. 
    The optimization information from Synopsys IC-Compiler II ($ICC2$) \cite{icc2} is utilized in training to guide the gradients of our timing model.
    \item We perform gate sizing based on the size gradients of our timing model.
    A sizing-oriented straight-through estimator is developed to efficiently generate size gradients in discrete functions.
    An adaptive gradient back-propagation method is presented to update gate sizes effectively.
    \item Our framework is evaluated with open-source designs in TSMC 16nm technology. 
    The results demonstrate that it can achieve 16.29\%/18.61\% TNS/WNS improvements and 6.64$\times$ speedup on average compared with the commercial gate sizing tool $ICC2$.
\end{itemize}

\begin{figure*}[!t]
	\centerline{\includegraphics[width=0.7\textwidth]{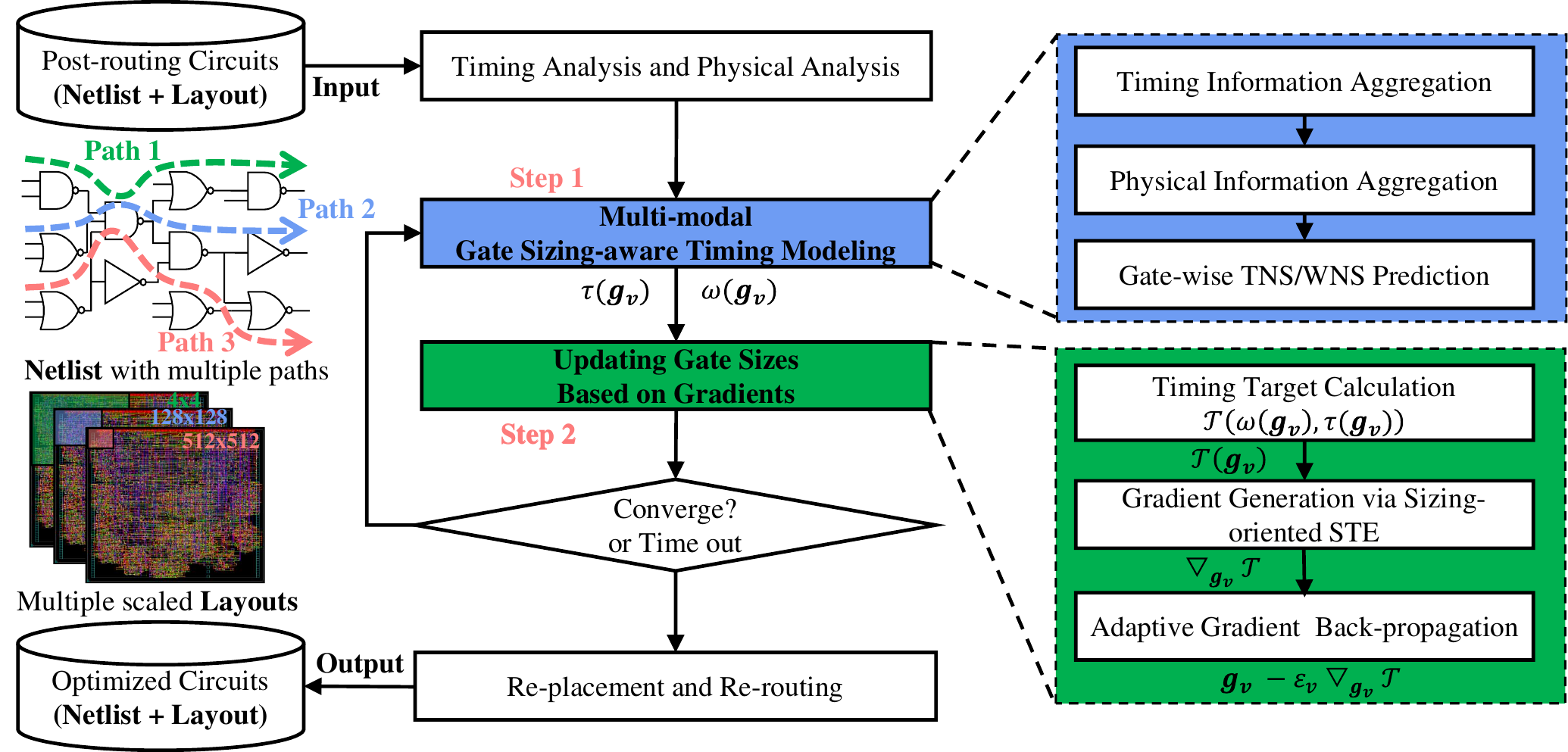}}
	\caption{The overall flow of our framework.}
	\label{fig:ourflow}
\end{figure*}
\section{Preliminaries}
\subsection{Timing optimization}
Timing optimization is important in the circuit design flow to fix timing issues on timing paths.
In circuits, timing paths are composed of a startpoint and an endpoint. 
The startpoint is a primary input or a register’s output pin, while the endpoint is a primary output or a register’s input pin. 
And the path slacks of all paths are computed based on path delays and the target clock period.
Two metrics is used to evaluate timing performance, including (1) the total negative slack (TNS),
which is the sum of the negative slacks observed at the primary outputs of the circuit; and (2) the worst negative slack (WNS),
which is the worst negative slack observed among all primary outputs of the circuit.
Timing optimization focus on improving timing performance through bringing changes to circuits. 
In our work, we look forward to achiving it through gate sizing. 
It is a representative technique \cite{kahng2011vlsi}.
It chooses a better size for each gate from the cell library to optimize overall timing performance.
Modern physically-aware gate sizing flows should not only consider timing information but also physical information, e.g. gate density and wire congestion, to avoid numerous iterations.

\subsection{Important Definitions}
We give some important definitions as follows:
\begin{mydefinition}[Gate-wise critical path]
    The most critical timing path through the target gate. 
\end{mydefinition}
\begin{mydefinition}[Gate-wise path group]
    The path group that is composed of critical timing paths through the target gate. 
\end{mydefinition}
\begin{mydefinition}[Gate-wise worst negative slack]
    The negative slack of gate-wise critical path for the target gate.
\end{mydefinition}
\begin{mydefinition}[Gate-wise total negative slack]
    The total negative slack of paths in the gate-wise path group for the target gate. 
\end{mydefinition}
\noindent \textit{Examples:} 
As shown in \Cref{fig:netlist}, Path 1, Path 2 and Path 3 are the gate-wise critical paths of gate $\operatorname{U4}$, $\operatorname{U7}$ and $\operatorname{U3}$, respectively.
For the gate $\operatorname{U4}$, Path 2 and Path 3 are included in the gate-wise path group for it.
The gate-wise worst negative slack of gate $\operatorname{U4}$ equals to -120ps.
The gate-wise total negative slack of gate $\operatorname{U4}$ equals to -285ps.

\subsection{Problem Formulation}
Based on these definitions, the problem of gate sizing can be formulated as:
\begin{myproblem}[Gate sizing]
Given a post-routing netlist with timing information on multiple critical timing paths and layout with physical information under multiple scales, our target is to achieve optimal gate sizes of all gates $\{\vec{g}_v$, $v \in \mathcal{V}\}$ based on the information to obtain optimized timing performance with smaller TNS and WNS, where $\mathcal{V}$ is the gate set.
\end{myproblem}

\section{Overall Flow}
As illustrated in \Cref{fig:ourflow}, we first briefly introduce the overall flow of our gate sizing framework.
The proposed framework can be divided into two steps: step 1 achieves the gate sizing-aware timing modeling based on multi-modal learning (\Cref{sec:model}) and step 2 updates gate size based on gradients (\Cref{sec:gradient}).
In step 1, we learn timing information on multiple paths through timing feature aggregation and physical information on multiple scaled layouts through physical feature aggregation jointly.
Based on learned information, we perform gate-wise TNS $\tau(\vec{g}_v)$ and WNS $\omega(\vec{g}_v)$ prediction where slack labels and gradient labels are used in the loss function to ensure high accuracy.
In step 2, we calculate the timing target $\mathcal{T}\left(\omega\left(\boldsymbol{g}_{\boldsymbol{v}}\right), \tau\left(\boldsymbol{g}_{\boldsymbol{v}}\right)\right)$ based on our timing model.
Then we generate timing target gradients $\nabla_{\vec{g_v}} \mathcal{T}$ w.r.t., gate sizes $\{\vec{g}_v$, $v \in \mathcal{V}\}$.
The sizing-oriented straight-through estimator helps to solve discrete issues. 
Finally, we update the gate size of each gate ($\vec{g_v} - \varepsilon_v \nabla_{\vec{g_v}} \mathcal{T}$) via the adaptive gradient backward propagation.
Our framework can optimize the timing performance of circuits, including TNS and WNS.
The details are discussed as follows.

\section{Gate Sizing-aware Timing Modeling}
\label{sec:model}
\begin{figure*}[!t]
    \centering
    \subfloat[Circuit netlist graph.]{\includegraphics[width=0.32\linewidth]{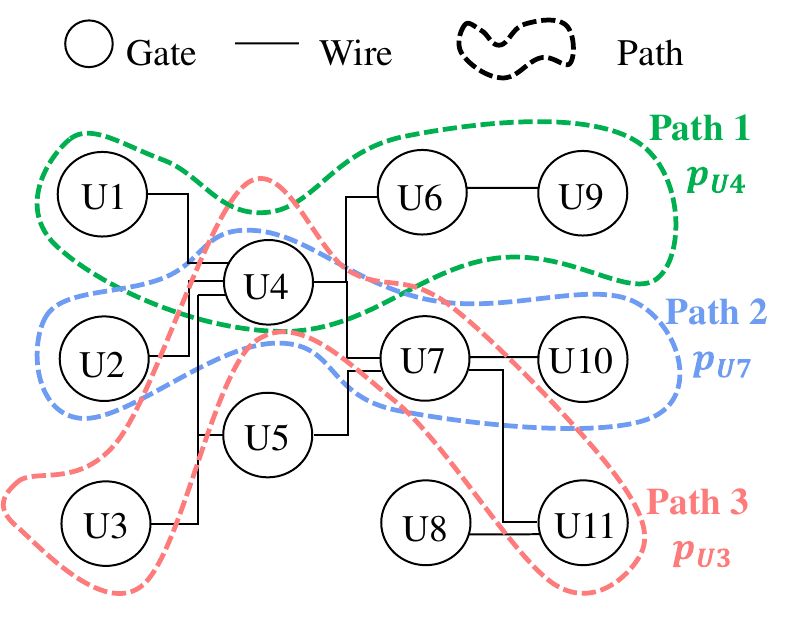} \label{fig:graph} }
    \subfloat[Physical features of multiple scaled layouts.]{\centering\includegraphics[width=0.32\linewidth]{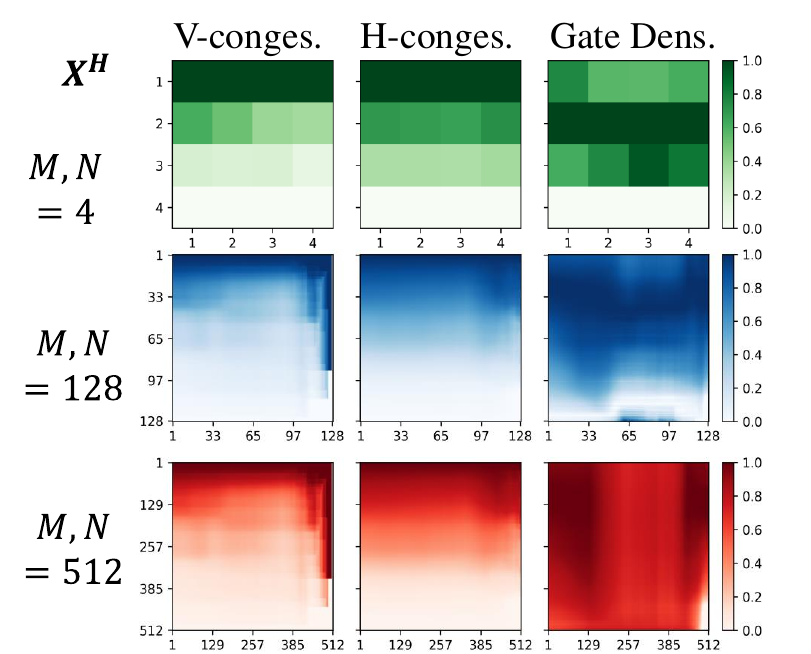} \label{fig:image2} }
    \subfloat[Gate-wise TNS labels and TNS gradient labels]{\centering\includegraphics[width=0.32\linewidth]{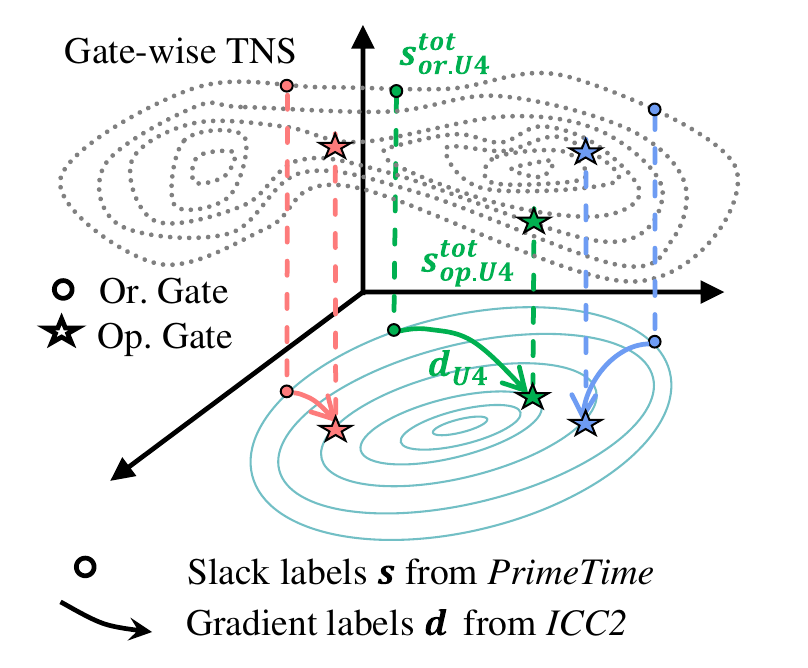} \label{fig:label} }
    \caption{Data representation in our work. ``Or. Gate'' and ``Op. Gate'' represent the original gate size and optimized gate size. } 
\end{figure*}

\subsection{Date Representation}
\noindent \textbf{Timing features on netlists: }
As shown in \Cref{fig:graph}, we transfer the circuit netlist to a graph $\mathbb{G}=(\mathcal{V},\mathcal{E},\mathcal{P})$ consisting of a node set ($\mathcal{V}$), a edge set ($\mathcal{E}$) and a sub-graph set ($\mathcal{P}$).
Nodes are gates and edges are wires.
More importantly, sub-graphs are critical paths composed of gates and wires on paths.
The circuit graph $\mathbb{G}$ is represented with node feature matrix $\vec{X}^{T}$: $\{\vec{x}_v^{T}$, $v \in \mathcal{V}\}$, adjacency matrix $\vec{J}$.
The details of features in feature vector $\vec{x}_v^{T}$ includes: 
(1) Gate size $\{\vec{g}_v$, $v \in \mathcal{V}\}$: extracted by the cell type name, determining
the driving strength of the gate;
(2) Gate type: e.g., NAND, NOR, embedded as a one-hot vector;
(3) Wire capacitance and resistance: extracted from the SPEF files generated by StarRC \cite{starrc};
(4) Pin capacitance: extracted from the timing library.

\noindent \textbf{Physical features on layouts: }
We divide the overall layout into different scales with $M \times N$ grid cells.
In previous timing models \cite{wang2023restructure}, they work on layout under one scale.
Thus, the values of $M$ and $N$ are set to be constant, which equals $512$.
Achieving gate sizing based on physical information on different scaled layouts can obtain different results \cite{icc2}.
Thus, we collect physical features on multiple-scaled layouts where $M$ and $N$ are set to be different values.
Specifically, local and global physical information is collected on large-scale and small-scale layouts, respectively.
The detailed considered physical features should be closely correlated with gate sizing, which include:
(1) Vertical wire congestion;
(2) Horizontal wire congestion;
(3) Gate density.
\Cref{fig:image2} gives examples of physical features $\vec{X}^{H}$: $\{\vec{x}_v^{H}$, $v \in \mathcal{V}\}$ on different scaled layouts for one Opencore design $\operatorname{NOVA}$.

\noindent \textbf{Slack labels and Gradient labels: }
As shown in \Cref{fig:label}, there are two kinds of labels used for training our timing model, including slack labels and gradient labels.
Slack labels $\{{s}_v^{tot}$, ${s}_v^{wst}$; $v \in \mathcal{V}\}$ are gate-wise total slacks and gate-wise worst negative slacks for all gates, which are generated via static timing analysis based on Synopsys $PrimeTime$ \cite{primetime};
Different from previous works, the gradient labels $\{{d}_v^{tot}$ and ${d}_v^{wst}$, $v \in \mathcal{V}\}$ are the gate sizing directions to optimize TNS and WNS, which are generated based on gate sizing results of $ICC2$. 
For gate $v$, given the original and $ICC2$ optimized gate sizes $\{\vec{g}_{or.v}, \vec{g}_{op.v}\}$ and gate-wise TNS results $\{s_{or.v}^{tot}, s_{op.v}^{tot}\}$, ${d}_v^{tot}$ can be computed as: $d_{v}^{tot}$ = $(s_{op.v}^{tot}-s_{or.v}^{tot})/(\vec{g}_{op.v}-\vec{g}_{or.v})$.
In the same way, $d_{v}^{wst}$ is obtained.
Our work focuses on achieving timing optimization by gate sizing.
The optimization direction of $ICC2$ is the best possible after many explorations.
Thus, the gradient labels can help speed up the optimization process and avoid local optimal problems.
\subsection{Timing Feature Aggregation}
\label{sec:timing}
\begin{figure}[tb!]
	\centerline{\includegraphics[width=0.50\textwidth]{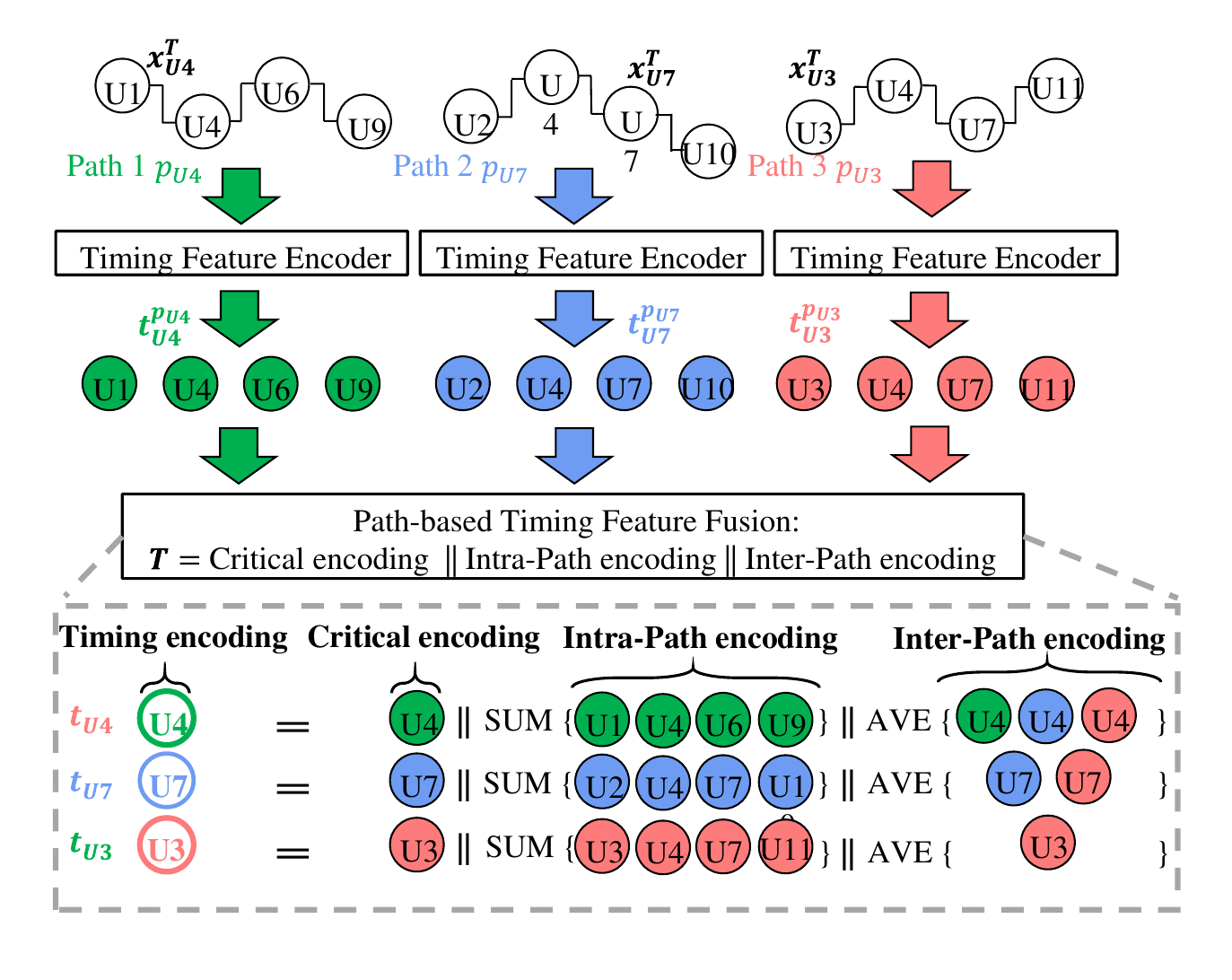}}
	\caption{An example of timing information aggregation on multiple paths.}
	\label{fig:pathlearn}
\end{figure}

Gate sizing for one target gate should consider all critical paths through it to achieve timing optimization and degradation trade-off.
One example of the timing information aggregation flow for gate $\operatorname{U4}$, gate $\operatorname{U7}$ and gate $\operatorname{U3}$ is shown in \Cref{fig:pathlearn}.
Given original timing features $\vec{X}^{T}$: $\{\vec{x}_v^{T}$, $v \in \mathcal{V}\}$ as input, the flow outputs path aggregated timing features $\vec{T}$: $\{\vec{t}_v$, $v \in \mathcal{V}\}$.
For one gate $v$, the gate-wise critical path and path group of gate $v$ are $p_v$ and $\mathcal{P}_v$, respectively.
The detailed progress in generating $\vec{t}_v$ is discussed as follows.

\noindent \textbf{Timing feature encoder:} Given the input timing features $\{\vec{x}_v^{T}$, $v \in \mathcal{V}\}$, Transformers achieve timing feature encoding path by path and output $\{\vec{t}_v^{p}$, $v \in \mathcal{V}$, $p \in \mathcal{P}\}$. On path $p$, the $\vec{t}_v^{p}$ is generated via:
\begin{equation}
\vec{t}_{v}^{p} = \text{Transformer}(\{\vec{x}_{u}^{T},  u \in \mathcal{N}_p\}, \vec{x}_{v}^{T}),
\end{equation}
where $\mathcal{N}_{p}$ is the gate set of path $p$.
On gate-wise critical path $p_v$, the results of timing feature encoding $t_v^{p_v}$ is regarded as the critical encoding of gate $v$.
Here, Transformer proposed in TransSizer \cite{nath2022transsizer} is used in our work.
This part can collect timing information in a path-by-path way.

\noindent \textbf{Path-based timing feature fusion:}
In the timing feature encoder, similar to TransSizer \cite{nath2022transsizer}, the timing feature on paths is learned path by path.
However, for real optimal gate sizing, timing performances on all critical timing paths through one gate should be considered jointly to achieve timing optimization and degradation trade-off.
In this work, we focus on achieving timing information aggregation on multiple timing paths.
The final path aggregated timing feature $\vec{t}_v$ of gate $v$ is composed of three parts, including critical encoding, intra-path encoding and inter-path encoding.

(1) In the critical encoding part, we obtain the timing feature encoding result $\vec{t}_v^{p_v}$ of gate $v$ on its gate-wise critical path $p_v$.
The slack of $p_v$ is gate-wise WNS of $v$ and is dominant in gate-wise TNS.
The influence of gate-sizing happened on gate $v$ on path $p_v$ is modeled accurately in this part. 
Thus, it helps improve the prediction accuracy of gate-wise WNS and TNS efficiently.
(2) In the intra-path encoding part, we obtain the timing feature encoding results of gate-wise critical path $p_v$ via pooling all gates' timing feature encoding results on it.
It helps our timing model to capture the relationship between gate $v$ and other gates on $p_v$ for accurate gate-wise WNS and TNS predictions.
(3) In the inter-path encoding part, we combine the timing feature encoding results of gate $v$ on all critical paths in gate-wise path group $\mathcal{P}_v$ through average pooling.
This part captures the relationships between gate $v$ and all critical paths through it.
It achieves accurately modeling the timing variations on paths in $\mathcal{P}_v$ caused by gate sizing on gate $v$.  
It is helpful to achieve accurate gate-wise TNS prediction.
The path aggregated timing feature $\vec{t}_v$ is computed as:
\begin{equation}
\begin{aligned}
& \vec{t}_{v} = \underbrace{(\vec{t}_v^{p_v}, p_v \  \text{is critical path})}_{\text{Critical Encoding}} \quad \\
& || \underbrace{{\operatorname{SUM}} \  ({t}_{u_v}^{p_v}, {u_v \in \mathcal{N}_{p_v}} )}_{\text{Intra-path Encoding}}
|| \underbrace{\operatorname{AVE}\ ({t}_{v}^{p}, {p \in \mathcal{P}_{v}} )}_{\text{Inter-path Encoding}},
\end{aligned}
\end{equation}
where $p_v$ is the gate-wise critical path of gate $v$, and $\mathcal{N}_{p_v}$ is the gate set of path $p_v$.
$\mathcal{P}_v$ is the gate-wise path group of gate $v$.
$\operatorname{SUM}$ and $\operatorname{AVE}$ represent sum pooling and average pooling operations, respectively.
\subsection{Physical Feature Aggregation}
\label{sec:physical}
\begin{figure}[tb!]
	\centerline{\includegraphics[width=0.50\textwidth]{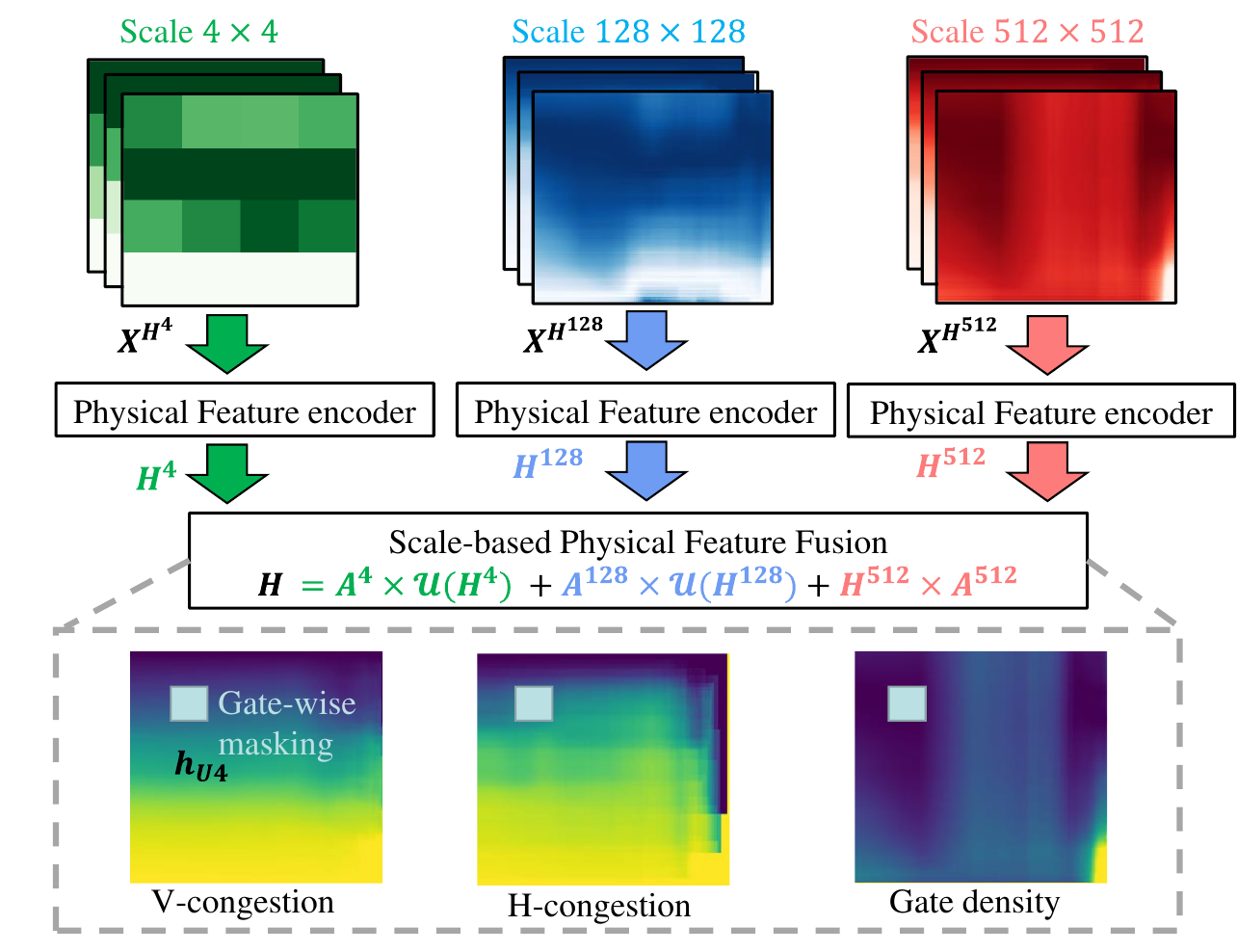}}
	\caption{An example of physical information aggregation on multiple scaled layouts.}
	\label{fig:scalelearn}
\end{figure}

Aggregating the differentiated information on different scales benefits capturing the circuit timing variation caused by re-placement and re-routing after gate sizing.
One example of the physical information aggregation flow for layouts under $4\times4, 128\times128$ and $512\times 512$ scales is illustrated in the right part of \Cref{fig:scalelearn}.
Given the input physical feature $\vec{X}^{{H}^{M \times N}}$, the flow outputs the scale aggregated physical feature $\vec{H}$ through combining information on different-scaled layouts.
Since $M$ equals to $N$ in this example, $\vec{X}^{{H}^{M \times N}}$ and $\vec{H}^{M \times N}$ can be represented with $\vec{X}^{{H}^{M}}$ and $\vec{H}^{M}$.
The flow is divided into two modules: the physical feature encoder module and the scale-based physical feature fusion module.
The flow captures global and local physical information jointly on layouts.

\noindent \textbf{Physical feature encoder:} 
We start by encoding physical features under different scales independently and generate $\vec{H}^{M \times N}$.
For the trade-off between efficiency and effectiveness,  the ResNet \cite{he2016deep} and ASPP \cite{azad2020attention} are used to extract and compress input physical features, respectively.
Specially, ResNet layer is constructed based on the feature extraction part of ResNet-50 without all other necessary parts.
The ASPP layer is composed of five ``Conv-BN-ReLU" branches. 
The kernel sizes and dilation rates of them are 1; 3; 3; 3; 1 and 1; 2; 5; 7; 1. 
All convolution operations use the padding to ensure that the input and output sizes are consistent. 
A global average pooling operation and an up-sampling operation are used before and after the second branch to capture the global physical information and restore it to the original size. 
All results of the five branches are concatenated along the channel dimension and fused by a branch to obtain the output.
Thus, $\vec{H}^{M \times N}$ can be computed as:
\begin{equation}
\vec{H}^{M \times N}  = \text{ResNet-ASPP}(\vec{X}^{{H}^{M \times N}}).
\end{equation}
Next, these features are fed successively to the scale-based physical feature fusion module for subsequent processing.

\noindent \textbf{Scale-based physical feature fusion:} 
We set one main scale, which equals $512 \times 512$, the biggest scale we selected in our work.
For physical features on the small-scaled layout, we directly up-sample them by the bi-linear interpolation.
Based on all encoded physical features from multiple scaled layouts $\{\vec{H}{^{1 \times 1}}, \vec{H}{^{2 \times 2}}, \dots, \vec{H}^{256 \times 256}, \vec{H}^{512 \times 512}\}$, the scale attention $\vec{A}$ corresponding to each scale can be obtained.
The process is formulated as:
\begin{equation}
\vec{A} = \sigma(\Psi\{\mathcal{U}(\vec{H}{^{1 \times 1}}) || \dots || \vec{H}^{512 \times 512}\}),
\end{equation}
where $\Psi$ indicates the stacked ``Conv-BN-ReLU" layers which are commonly used in convolutional neural networks \cite{he2016deep}.
$||$ represents the concatenation operations.
$\mathcal{U}(\cdot)$ refers to the bi-linear interpolation operations for up-sampling mentioned above.
$\sigma$ is $\operatorname{Softmax}$ activation operation in our work.
Based on generated scale attention, we can obtain the final scale aggregated physical feature $\vec{H}$ by combining the scale-specific information jointly. 
Inspired by \cite{wang2023restructure}, gate-wise masking can help us to get scale aggregated physical feature for each gate $\{\vec{h}_v$, $v \in \mathcal{V}\}$. They can be computed as :
\begin{equation}
\vec{H} = \sum_{\text{all scales}}\  \vec{A}^{M \times N} \times \mathcal{U}(\vec{H}^{M \times N}),\ \ \vec{h}_v = \vec{M}_v \vec{H},
\end{equation}
where $\mathcal{U}(\cdot)$ is unnecessary for the features on main scale $\vec{H}^{512 \times 512}$.
$\vec{M}_v$ is the gate-wise mask for gate $v$.
These designs can selectively aggregate the scale-specific physical features to explore subtle but critical information among different scales.
It helps predict TNS and WNS improvements and degradations induced by gate sizing after re-placement and re-routing.

\subsection{Gate-wise TNS and WNS Prediction}
Based on the path aggregated timing features $\vec{T}$: $\{\vec{t}_v, v \in \mathcal{V}\}$ and scale aggregated physical features $\vec{H}$: $\{\vec{h}_v, v \in \mathcal{V}\}$, we use multilayer perceptron layers $MLP^{\tau}$ and $MLP^{\omega}$ to predict the gate-wise total negative slacks and gate-wise worst negative slacks for all gates. 
\begin{equation}
    \tau(\vec{g}_v) = MLP^{\tau}(\vec{t}_v || \vec{h}_v), \ \ \ \omega(\vec{g}_v) = MLP^{\omega}(\vec{t}_v || \vec{h}_v),
\end{equation}
where $\vec{g}_v$ is the gate size of gate $v$.
Both the slack labels and gradient labels are used in loss functions of $\tau(\vec{g}_v)$ and $\omega(\vec{g}_v)$.
The slack labels play important and fundamental roles in improving timing model accuracy.
The gradient labels can be regarded as constraints to guide optimization directions.
Combining these two labels, the loss functions used for training are illustrated in \Cref{equ:loss}.
\begin{equation}
\begin{aligned}
    &\mathcal{L}^{\tau}=\sum_{v \in \mathcal{V}}\{\underbrace{({s}_{v}^{tot}-\tau(\vec{g}_{v}))^2}_{\text{slack labels}} + \underbrace{({d}_v^{tot}-\nabla_{\vec{g_v}}\tau)^2}_{\text{gradient labels}}\}, \\
    &\mathcal{L}^{\omega}=\sum_{v \in \mathcal{V}}\{\underbrace{({s}_{v}^{wst}-\omega(\vec{g}_{v}))^2}_{\text{slack labels}} + \underbrace{({d}_v^{wst}-\nabla_{\vec{g_v}}\omega)^2}_{\text{gradient labels}}\},
    \label{equ:loss}
\end{aligned}
\end{equation}
where ${s}_v^{tot}$ and ${s}_v^{wst}$ are gate-wise total slacks and gate-wise worst negative slacks generated via Synopsys $PrimeTime$; 
${d}_v^{tot}$ and ${d}_v^{wst}$ are gradients of gate-wise total slacks and gate-wise worst negative slacks generated after Synopsys $ICC2$ gate sizing;
$\nabla_{\vec{g_v}}\tau$ and $\nabla_{\vec{g_v}}\omega$ are gradients of $\tau(\vec{g}_v)$ and $\omega(\vec{g}_v)$ w.r.t.. gate size, where the detailed flow to generate them is discussed in \Cref{sec:grdgen}.

\section{Updating Gate Size Based on Gradients}
\label{sec:gradient}
\begin{figure}[tb!]
	\centerline{\includegraphics[width=0.5\textwidth]{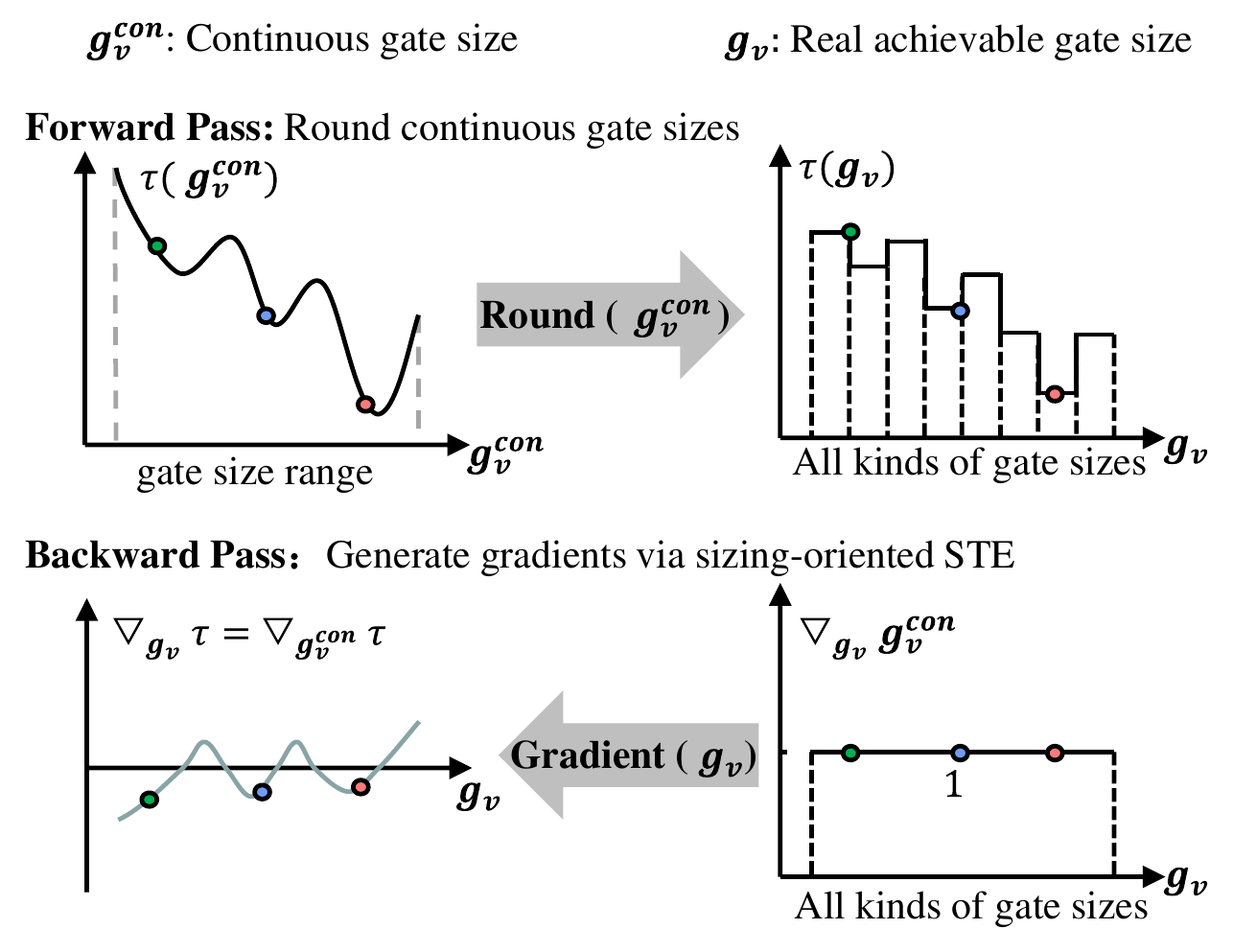}}
	\caption{An example flow of generating gradients of gate-wise TNS w.r.t. gate sizes via the sizing-oriented straight-through estimator (STE).}
	\label{fig:ste}
\end{figure}

\subsection{Timing Target Calculation}
\label{sec:target}
After obtaining the well-trained $\tau(\vec{g}_v)$ and $\omega(\vec{g}_v)$, the timing target $\mathcal{T}(\vec{g}_v)$ for gate sizing is calculated based on predicted gate-wise TNS and WNS results.
Different from previous work \cite{pham2023agd}, we consider all critical paths in the timing target $\mathcal{T}(\vec{g}_v)$ rather than the worst path on each point.
This is because focusing on optimizing the worst path on each point might cause timing degradations on other critical paths on the same point.
It makes many time-consuming iterations and causes local optimal problems in other works \cite{nath2022transsizer}.
The timing target $\mathcal{T}(\vec{g}_v)$ can computed as:
\begin{equation}
\mathcal{T}(\{\vec{g}_v, v\in \mathcal{V}\}) = 
\underbrace{\frac{\mu^{\tau}}{N}\sum_{v \in \mathcal{V}} \min \{0, \tau(\vec{g}_v)\}}_{\text{TNS Target}} + \underbrace{\mu^{\omega} \vphantom{\sum_{v \in \mathcal{V}}}\min_{v \in \mathcal{V}} \omega(\vec{g}_v)}_{\text{WNS Target}},
\label{equ:target}
 \end{equation}
where $\mu^{\tau}$ and $\mu^{\omega}$ are weights for the TNS target and WNS target, respectively.
$N$ is the number of gates with negative gate-wise total negative slacks.
As the WNS target and TNS target contain minimum operation, directly applying the timing target $\mathcal{T}$ for backward propagation leads to a cut-off in some timing paths. 
To overcome the drawback, we follow the method proposed in \cite{liu2023concurrent} to smooth the minimum and maximum operations.
In details, these operations are replaced with the $\operatorname{Log-Sum-Exp}$ function as follows,
\begin{equation}
\label{equ:lse}
	L S E\left(\omega(\vec{g}_v), v\in \mathcal{V}\right)=\gamma \log \left(\sum_{v\in \mathcal{V}} \exp \frac{\vec{g}_v}{\gamma}\right),
\end{equation}
where $\gamma$ is the critical parameter to adjust the degree of smoothing where a larger $\gamma$ causes smoother results with lower approximation accuracy. 
Similarly, the minimum operation is smoothed by the inverse values.
Thus, the value of $\gamma$ plays an important role in our work to achieve efficient timing optimization and is necessary to be selected carefully.
After that, we can get the smoothed $\mathcal{T}(\vec{g}_v)$.
Based on smoothed $\mathcal{T}(\vec{g}_v)$, the timing optimization gradients w.r.t.~gate size ($\nabla_{\vec{g_v}} \mathcal{T}$) can be computed automatically via backward propagation, which can be used in our gate-sizing framework.
\subsection{Gradient Generation}
\label{sec:grdgen}
As shown in \Cref{equ:target}, the first and fundamental task is to calculate gradients of our timing model w.r.t. gate sizes ($\nabla_{\vec{g_v}} \tau$ and $\nabla_{\vec{g_v}} \omega$) before generating gradients of timing target $\nabla_{\vec{g_v}} \mathcal{T}$.
Since gate size $\vec{g}_v$ of each gate is discrete rather than continuous, the $\operatorname{round}$ operation is a necessity in our timing models $\tau(\vec{g}_v)$ and $\omega(\vec{g}_v)$ during forward pass.
As illustrated in \Cref{fig:ste}, the continuous gate size $\vec{g}_v^{con}$ can be translated into real achievable gate size $\vec{g}_v$ after it.
However, it makes our timing models not differentiable w.r.t. gate sizes.
Thus, the sizing-oriented straight-through estimator is developed to solve the issue and generate gradients ($\nabla_{\vec{g_v}} \tau$ and $\nabla_{\vec{g_v}} \omega$) accurately for gate sizing.
\Cref{fig:ste} gives an example of generating $\nabla_{\vec{g_v}} \tau$ via the sizing-oriented straight-through estimator. 
In it, the gradient of the $\operatorname{round}$ operator is approximated as 1.
Based on the approximation, we can get the $\nabla_{\vec{g_v}} \tau$ as:
\begin{equation}
	\begin{aligned}
& \nabla_{\vec{g}_v} \vec{g}_v^{con} = 1 \rightarrow
\nabla_{\vec{g}_v} \tau = \nabla_{\vec{g}_v^{con}} \tau, \\ 
& \nabla_{\vec{g}_v} \omega = \nabla_{\vec{g}_v^{con}} \omega, \quad v \in \mathcal{V}.
\end{aligned}
\end{equation}

This simple approximation function works well in quantization-aware training works.
Fortunately, it is also a good method to solve discrete issues in gate sizing-aware timing models.
We give an explanation for the efficiency as follows:
In quantization works, the float point variables are quantized with bit-wise variables.
Similar to quantization works, the gate size can be continuous while designing.
It is quantized while generating standard libraries to compact the library size and improve design efficiency \cite{kahng2011vlsi}.
Thus, discrete issues in gate-sizing work are the same as quantization works.
As shown in \Cref{fig:ste}, the relationship between different sizes can be retained in our work.
Based on the generated $\nabla_{\vec{g_v}} \tau$ and $\nabla_{\vec{g_v}} \omega$, the timing targets gradients w.r.t.~gate sizes $\nabla_{\vec{g_v}} \mathcal{T}$ can be computed automatically and accurately before backward propagation.
Only the feature of gate size as ‘gradient required’.

\subsection{Adaptive Gradient Back-propagation}
\label{sec:propa}
After obtaining timing target gradients w.r.t.. gate sizes $\nabla_{\vec{g_v}} \mathcal{T}$, the stochastic optimization algorithm proposed in Adam \cite{kingma2014adam} can be applied to optimize the timing target $\mathcal{T}$ via gradient back-propagation.
The gate size in our work can be updated as:
\begin{equation}
~\label{eq-vanillagradient}
\vec{g}_v := \vec{g}_v - \varepsilon_v \nabla_{\vec{g}_v} \mathcal{T}, \quad v \in \mathcal{V},
\end{equation}
where $\varepsilon_v$ is the learning rate in Adam.
However, if we directly perform gradient descent following \Cref{eq-vanillagradient}, it is difficult to solve the gate sizing problem with high efficacy on large-scale circuits with many gates.
The problem is caused by the high-dimensional issue.
When multiple gate sizes change simultaneously, there may be variations in gradient estimation. 
This is because there are interdependencies among different gates. 
In the experience of physical designers, it is a common practice to fix some gates while performing gate sizing on others for achieve timing optimization.

In our work, we incorporate the experience of physical designers and use an adaptive learning rate $\varepsilon_v$ to update gate sizes based on gradients. 
If we employ an alternating optimization scheme with sampling, it may result in unacceptable runtime costs. 
Instead, we utilize the well-known technique of Gumbel-Softmax \cite{jang2016categorical} to achieve adaptive back-propagation via sampling:
\begin{equation}
\varepsilon_v=\frac{\exp \left(\left(\log \left(\omega(\vec{g}_v)\right)+n_v\right) / \lambda\right)}{\sum_{i \in \mathcal{V}} \exp \left(\left(\log \left(\omega(\vec{g}_i)\right)+n_i\right) / \lambda\right)}, \quad v \in \mathcal{V},
\label{equ:g-softmax}
\end{equation}
where $n_v$ and $n_i$ are independent and identically distributed samples drawn from Gumbel distribution.
$\lambda$ represents the temperature parameter.
Our intuition is that since timing issues are determined by their worst-case scenario, we use the normalized result of gate-wise WNS $\omega(\vec{g}_v)$ as the probability value for sampling. 
For gates with larger gate-wise WNS values, which are bottlenecks in timing, more probability is allocated for gradient sampling and gradient back-propagation.
It means they should be solved with higher priority.

\begin{table*}[!t]

	\centering
    \footnotesize
	\caption{Benchmark statistics. The units of ``WNS'' and ``TNS'' are $ns$. And the unit of ``POW'' is $mW$.} 
	\label{tab:0}
	{
		\begin{tabular}{|c|c|c|c|c|c|c|c|}
			\hline
			Circuit   & \#gates & \#wires & \#CPs  & WNS     & TNS        & NVE        \\ \hline \hline
			DMX       & 11616   & 11671   & 7068   & -0.4112 & -2.9417    & 170     \\
			GFX       & 11004   & 11156   & 18011  & -0.6688 & -107.1956  & 1042    \\
			AC97      & 4954    & 5046    & 3102   & -0.2281 & -7.7943    & 74      \\
			VGA       & 32727   & 32863   & 10860  & -0.8694 & -237.1859  & 2852    \\
			NOVA      & 105756  & 107641  & 64325  & -1.6939 & -1982.8487 & 12007  \\ \hline
			Tot.Train & 166057  & 168377  & 103366 & -3.8714 & -2337.9662 & 16145 \\ \hline
			TOP   & 3943    & 4121    & 5660   & -0.2500 & -16.7637   & 216     \\
			ECG       & 39992   & 40941   & 40963  & -0.5941 & -771.1540  & 3830   \\
			ETH       & 25327   & 25450   & 11681  & -1.4086 & -401.2948  & 924     \\
			USB       & 6666    & 7000    & 6143   & -0.4830 & -26.1364   & 93      \\
			TATE      & 153192  & 154720  & 73147  & -1.5241 & -2149.5106 & 5272   \\ \hline
			Tot.Test  & 229120  & 232232  & 137594 & -4.2598 & -3364.8595 & 10335  \\ \hline
		\end{tabular}
	}
\end{table*}

\begin{table*}[t]
	\centering
    \footnotesize
	\caption{Gate-wise TNS $\tau(\vec{g}_v)$ and gate-wise WNS $\omega(\vec{g}_v)$ prediction accuracy. The unit of ``$MAE$'' is $ps$.}
	\label{tab:model}
{ 
	\begin{tabular}{|c|cc|cc|cc|cc|cc|cc|cc|cc|cc|cc|}
		\hline
		\multirow{3}{*}{Cir.} & \multicolumn{4}{c|}{DAC22 \cite{guo2022timing}}             & \multicolumn{4}{c|}{DAC23 \cite{wang2023restructure}}                
		& \multicolumn{4}{c|}{Ours Paths-only}              
		& \multicolumn{4}{c|}{Ours}             \\ \cline{2-17}
		& \multicolumn{2}{c|}{$\tau(\vec{g}_v)$} & \multicolumn{2}{c|}{$\omega(\vec{g}_v)$} & \multicolumn{2}{c|}{$\tau(\vec{g}_v)$} & \multicolumn{2}{c|}{$\omega(\vec{g}_v)$} & \multicolumn{2}{c|}{$\tau(\vec{g}_v)$} & \multicolumn{2}{c|}{$\omega(\vec{g}_v)$}	& \multicolumn{2}{c|}{$\tau(\vec{g}_v)$} & \multicolumn{2}{c|}{$\omega(\vec{g}_v)$} \\ \cline{2-17}
		& $R^2$  & $MAE$    & $R^2$  & $MAE$  & $R^2$  & $MAE$    & $R^2$  & $MAE$  & $R^2$  & $MAE$    & $R^2$  & $MAE$  & $R^2$  & $MAE$   & $R^2$  & $MAE$   \\ \hline \hline
		DMX                     & 0.75 & 44.35  & 0.76    & 36.14  & 0.82 & 38.13  & 0.87    & 27.75    & 0.85 & 14.28  & 0.87    & 11.29  & \textbf{0.96} & \textbf{6.29}  & \textbf{0.95}    & \textbf{3.98}  \\
		GFX                      & 0.71 & 36.21 & 0.74   & 29.34  & 0.76 & 28.15 & 0.81   & 22.98    & 0.82 & 12.54 & 0.86   & 10.89  & \textbf{0.96} & \textbf{5.42} & \textbf{0.98}    & \textbf{3.11}  \\
		AC97                    & 0.78 & 47.56  & 0. 8    & 32.68  & 0.84 & 35.24  & 0.86     & 27.77   & 0.84 & 19.12  & 0.86     & 16.24  & \textbf{0.94} & \textbf{7.88}  & \textbf{0.97}     & \textbf{5.06}  \\
		VGA                      & 0.73 & 56.13  & 0.78   & 42.97  & 0.77 & 45.23  & 0.82    & 36.88    & 0.86 & 17.88  & 0.89    & 14.25  & \textbf{0.94} & \textbf{6.92}  & \textbf{0.97}    & \textbf{3.78}  \\
		NOVA                   & 0.79 & 82.14 & 0.81  & 71.54  & 0.82 & 50.13 & 0.84    & 46.37    & 0.86 & 37.28 & 0.90   & 23.99  & \textbf{0.95} & \textbf{7.99} & \textbf{0.98}    & \textbf{4.28}  \\ \hline
  		Ave.                & 0.75  & 53.28   & 0.78 & 42.54  & 0.81   & 39.38 & 0.84  & 34.43    & 0.84   & 20.22 & 0.88  & 15.33 & \textbf{0.95}  & \textbf{6.50}  & \textbf{0.97}     & \textbf{4.04} \\ \hline \hline
		TOP              & 0.67 & 24.24  & 0.71     & 19.15  & 0.75 & 17.21   & 0.78    & 13.99    & 0.79 & 13.75  & 0.83     & 8.98  & \textbf{0.94} & \textbf{5.23}  & \textbf{0.97}     & \textbf{2.35}   \\
		ECG                      & 0.58 & 37.26   & 0.64    & 27.82  & 0.71 & 29.12   & 0.76    & 27.14    & 0.79 & 14.23  & 0.82    & 12.59  & \textbf{0.90}  & \textbf{4.29}  & \textbf{0.94}     & \textbf{3.20}  \\
		ETH                      & 0.69 & 42.75  & 0.72    & 34.62  & 0.77 & 35.14  & 0.81    & 26.32   & 0.83 & 11.27  & 0.87    & 12.17  & \textbf{0.93} & \textbf{3.97}  & \textbf{0.95}    & \textbf{2.12}  \\
		USB                      & 0.66 & 27.99  & 0.70    & 15.74  & 0.76 & 17.89  & 0.80    & 11.24  & 0.84 & 8.24  & 0.89    & 7.92  & \textbf{0.92} & \textbf{3.12}   & \textbf{0.94}    & \textbf{1.27}  \\
		TATE                     & 0.70 & 98.72  & 0.74    & 89.22 & 0.79 & 57.32  & 0.85    & 49.21  & 0.82 & 20.28  & 0.84    & 16.62 & \textbf{0.94} & \textbf{8.23}  & \textbf{0.96}     & \textbf{6.75} \\ \hline 
		Ave.                & 0.66  & 46.19   & 0.70 & 37.31  & 0.76  & 31.14   & 0.80 & 25.58  & 0.82  & 13.55   & 0.85 & 11.67  & \textbf{0.94}  & \textbf{4.77}  & \textbf{0.95}     & \textbf{3.14} \\ \hline 
	\end{tabular}
}
\end{table*}

\begin{table*}[t]
\centering
\footnotesize
\caption{Timing optimization result comparison between our framework and other gate sizing works.}
\label{tab:1}
\resizebox{1.02\linewidth}{!}
{ 
\begin{tabular}{|c|ccc|ccc|ccc|ccc|ccc|}
	\hline
	\multirow{2}{*}{Cir.} & \multicolumn{3}{c|}{$ICC2$}             & \multicolumn{3}{c|}{RL-Sizer \cite{lu2021rl}}         
	& \multicolumn{3}{c|}{TransSizer \cite{nath2022transsizer}}        
	& \multicolumn{3}{c|}{AGD \cite{pham2023agd}}              
	& \multicolumn{3}{c|}{Ours}             \\ \cline{2-16}
	& WNS   & TNS    & NVE   & WNS   & TNS    & NVE   & WNS   & TNS    & NVE   & WNS   & TNS    & NVE  & WNS   & TNS    & NVE   \\ \hline \hline
	DMX                      & -0.1596 & -0.9715  & 92      & -0.1465 & -0.8521  & 80      & -0.1632 & -1.2354  & 98      & -0.1508 & -0.9408  & 89      & \textbf{-0.1391} & \textbf{-0.7998}  & \textbf{72}      \\
	GFX                      & -0.4129 & -30.2528 & 150     & -0.3976 & -26.2132 & 129    & -0.4432 & -36.2589 & 192    & -0.4021 & -36.1854 & 138    & \textbf{-0.3659} & \textbf{-23.4321} & \textbf{98}     \\
	AC97                     & -0.0002 & -0.0003  & \textbf{2}       & -0.0003 & -0.0012  & 9       & -0.0004 & -0.0009  & 3     & -0.0003 & -0.0008  & 4       & \textbf{-0.0001} & \textbf{-0.0002}  & \textbf{2}       \\
	VGA                      & -0.0104 & -0.0456  & 14      & -0.0178 & -0.0829  & 47     & -0.0254 & -0.0618  & 32     & -0.0162 & -0.0745  & 42     & \textbf{-0.0093} & \textbf{-0.0408}  & \textbf{12}      \\
	NOVA                     & -0.5214 & -28.8674 & 72      & -0.4621 & -20.2178 & 59     & -0.5974 & -67.2348 & 348    & -0.5438 & -35.6218 & 103     & \textbf{-0.4339} & \textbf{-21.2486} & \textbf{31}      \\
	TOP                  & \textbf{-0.0002} & -0.0002  & \textbf{1}       & -0.0002 & -0.0008   & 4       & -0.0003 & -0.0011  & 8     & -0.0003 & -0.0013  & 9      & \textbf{-0.0002} & \textbf{-0.0002}  & \textbf{1}       \\
	ECG                      & -0.0012 & -0.0029  & 7       & -0.0023 & -0.0079   & 26      & -0.0017 & -0.0051   & 18      & -0.0019 & -0.0064  & 20    & \textbf{-0.0010}  & \textbf{-0.0022}  & \textbf{5}    \\
	ETH                      & -0.1596 & -0.9715  & 92      & -0.1465 & -0.8521  & 80     & -0.1632 & -1.2354  & 98    & -0.1508 & -0.9408  & 89      & \textbf{-0.1391} & \textbf{-0.7998}  & \textbf{72}     \\
	USB                      & -0.1199 & -4.5600  & 48     & -0.1052 & -4.0214  & 42    & -0.1256 & -5.2365  & 51      & -0.1201 & -4.5632  & 49    & \textbf{-0.1002} & \textbf{-3.514}   & \textbf{36}      \\
	TATE                     & -0.0013 & -0.0055  & 10     & -0.0021 & -0.0082  & 23     & -0.0018 & -0.0069  & 15    & -0.0019 & -0.0072  & 20     & \textbf{-0.0011} & \textbf{-0.0049}  & \textbf{8}      \\ \hline \hline
	Ave.                  & 1.0000  & 1.0000   & 1.00  & 1.1966  & 1.8236   & 2.22 & 1.4055  & 2.0088   & 2.52 & 1.2511  & 1.9684   & 2.42 & \textbf{0.8371}  & \textbf{0.8139}  & \textbf{0.78}     \\ \hline 
\end{tabular}
}
\end{table*}

\section{Experimental Results}
Our framework is implemented in Python with the Pytorch library and in C++.
The multi-modal timing model is trained on a Linux machine with 32 cores and 4 NVIDIA Tesla V100 GPUs. 
The training process takes about 4.5 hours using the parallel training method on 4 GPUs.
The total memory used is 128GB.
In timing target calculation, both the weights for TNS target $\mu^{\tau}$ and for WNS target  $\mu^{\omega}$ are set to 0.5.
And they can be adjusted to meet different timing requirements.
To smooth the penalty function described in \Cref{equ:lse}, we set $\gamma$ as 10.0.
The temperature parameter $\lambda$ used during adaptive gradient back-propagation equals 5.0.

In this work, we train our timing model and evaluate our framework using different open-source designs \cite{opencore}.
And our work can be applied to unseen design without re-training.
The benchmark circuits are synthesized with TSMC 16nm technology and details are shown in \Cref{tab:0}.
The circuit benchmarks are split into training and testing sets. 
The training and testing sets are determined by design scale in order to make balance. The timing
evaluation model is trained on the training set with a learning rate of 0.0004.
\#CPs represent the number of critical paths. 
WNS and TNS represent the worst and total negative slack of circuits. 
NVE represents the number of endpoints with timing violations. 
POW is the power consumption.
We compare our framework with the following advanced baselines: 
1) The commercial EDA tool $ICC2$;
2) RL-sizer \cite{lu2021rl};
3) Transizer \cite{nath2022transsizer};
4) AGD \cite{pham2023agd}: timing model proposed in \cite{guo2022timing}+gradient descent optimization.
\begin{figure*}[tb!]
	\centering
	\hspace{.1in}
	\subfloat[WNS improvement of TATE]{\pgfplotsset{
	width =0.50\textwidth,
	height=0.28\textwidth
}

\begin{tikzpicture}[scale=0.9]
	\begin{axis}[
		samples=1000,
		xmax=0.95, xmin=0.00,
		ymax=2.11, ymin=0.66,
		xtick={0.1,0.2,0.3,0.4,0.5,0.6,0.7,0.8,0.9},
		ytick={0.7,0.9,1.1,1.3,1.5,1.7,1.9,2.1},
		ymajorgrids=true,
		grid style=dashed,
		scaled ticks=false,
		legend pos=north east,
		xticklabel style={/pgf/number format/fixed},
		yticklabel style={/pgf/number format/.cd, fixed, fixed zerofill, precision=1, /tikz/.cd},
		xlabel={$\mu^{\omega}$ ($\mu^{\tau}$=$1-\mu^{\omega}$)},
		ylabel={Normalized WNS},
		xlabel near ticks,
		ylabel near ticks,
		legend style={
			draw=none,
			at={(0.5,1.0)},
			anchor=south,
			legend columns=-1,
		}
		]
		
		\pgfplotstableset{
			create on use/x rel/.style={
				create col/expr={
					\thisrow{0}
				}
			},
			create on use/y rel/.style={
				create col/expr={
					\thisrow{1}
				}
			}
		}

		\addplot +[myblue1, line width=1pt, mark=star, mark options={scale=1.0, fill=red},text mark as node=true ] table [x ={er}, y={cpd}] {wnstaterl.dat};
		\addplot +[mygreen2, line width=1pt, mark=o, mark options={scale=0.5, fill=myorange},text mark as node=true ] table [x ={er}, y={cpd}] {wnstateagd.dat};
		\addplot +[myred1, line width=1pt, mark=square, mark options={scale=0.5, fill=mymiddle},text mark as node=true] table [x ={er}, y={cpd}] {wnstateours.dat};
		\legend{RL-Sizer,AGD,Ours}
		
	\end{axis}

\end{tikzpicture} 
		\label{fig:wnstate}}
	\hspace{.1in}
	\subfloat[TNS improvement of TATE]{\pgfplotsset{
	width =0.50\textwidth,
	height=0.28\textwidth
}

\begin{tikzpicture}[scale=0.9]
	\begin{axis}[
		samples=1000,
		xmax=0.95, xmin=0.00,
ymax=2.11, ymin=0.66,
xtick={0.1,0.2,0.3,0.4,0.5,0.6,0.7,0.8,0.9},
ytick={0.7,0.9,1.1,1.3,1.5,1.7,1.9,2.1},
		ymajorgrids=true,
		grid style=dashed,
		scaled ticks=false,
		legend pos=north east,
		xticklabel style={/pgf/number format/fixed},
		yticklabel style={/pgf/number format/.cd, fixed, fixed zerofill, precision=1, /tikz/.cd},
		xlabel={$\mu^{\tau}$ ($\mu^{\omega}$=$1-\mu^{\tau}$)},
		ylabel={Normalized TNS},
		xlabel near ticks,
		ylabel near ticks,
		legend style={
	draw=none,
	at={(0.5,1.0)},
	anchor=south,
	legend columns=-1,
}
		]
		
		\pgfplotstableset{
			create on use/x rel/.style={
				create col/expr={
					\thisrow{0}
				}
			},
			create on use/y rel/.style={
				create col/expr={
					\thisrow{1}
				}
			}
		}

		\addplot +[myblue1, line width=1pt, mark=star, mark options={scale=1.0, fill=red},text mark as node=true ] table [x ={er}, y={cpd}] {tnstaterl.dat};
\addplot +[mygreen2, line width=1pt, mark=o, mark options={scale=0.5, fill=myorange},text mark as node=true ] table [x ={er}, y={cpd}] {tnstateagd.dat};
\addplot +[myred1, line width=1pt, mark=square, mark options={scale=0.5, fill=mymiddle},text mark as node=true] table [x ={er}, y={cpd}] {tnstateours.dat};
		\legend{RL-Sizer,AGD,Ours}

	\end{axis}

\end{tikzpicture} \label{fig:tnstate}}	\\
	\hspace{.1in}
	\subfloat[WNS improvement of ECG]{\pgfplotsset{
	width =0.50\textwidth,
	height=0.28\textwidth
}

\begin{tikzpicture}[scale=0.9]
	\begin{axis}[
		samples=1000,
		xmax=0.95, xmin=0.00,
		ymax=2.11, ymin=0.66,
		xtick={0.1,0.2,0.3,0.4,0.5,0.6,0.7,0.8,0.9},
		ytick={0.7,0.9,1.1,1.3,1.5,1.7,1.9,2.1},
		ymajorgrids=true,
		grid style=dashed,
		scaled ticks=false,
		legend pos=north east,
		xticklabel style={/pgf/number format/fixed},
		yticklabel style={/pgf/number format/.cd, fixed, fixed zerofill, precision=1, /tikz/.cd},
		xlabel={$\mu^{\omega}$ ($\mu^{\tau}$=$1-\mu^{\omega}$)},
		ylabel={Normalized WNS},
		xlabel near ticks,
		ylabel near ticks,
		legend style={
			draw=none,
			at={(0.5,1.0)},
			anchor=south,
			legend columns=-1,
		}
		]
		
		\pgfplotstableset{
			create on use/x rel/.style={
				create col/expr={
					\thisrow{0}
				}
			},
			create on use/y rel/.style={
				create col/expr={
					\thisrow{1}
				}
			}
		}

		\addplot +[myblue1, line width=1pt, mark=star, mark options={scale=1.0, fill=red},text mark as node=true ] table [x ={er}, y={cpd}] {wnsecgrl.dat};
		\addplot +[mygreen2, line width=1pt, mark=o, mark options={scale=0.5, fill=myorange},text mark as node=true ] table [x ={er}, y={cpd}] {wnsecgagd.dat};
		\addplot +[myred1, line width=1pt, mark=square, mark options={scale=0.5, fill=mymiddle},text mark as node=true] table [x ={er}, y={cpd}] {wnsecgours.dat};
		\legend{RL-Sizer,AGD,Ours}
		
	\end{axis}

\end{tikzpicture} 
		\label{fig:wnsecg}}
	\hspace{.1in}
	\subfloat[TNS improvement of ECG]{\pgfplotsset{
	width =0.50\textwidth,
	height=0.28\textwidth
}

\begin{tikzpicture}[scale=0.9]
	\begin{axis}[
		samples=1000,
		xmax=0.95, xmin=0.00,
		ymax=2.11, ymin=0.66,
		xtick={0.1,0.2,0.3,0.4,0.5,0.6,0.7,0.8,0.9},
		ytick={0.7,0.9,1.1,1.3,1.5,1.7,1.9,2.1},
		ymajorgrids=true,
		grid style=dashed,
		scaled ticks=false,
		legend pos=north east,
		xticklabel style={/pgf/number format/fixed},
		yticklabel style={/pgf/number format/.cd, fixed, fixed zerofill, precision=1, /tikz/.cd},
		xlabel={$\mu^{\tau}$ ($\mu^{\omega}$=$1-\mu^{\tau}$)},
		ylabel={Normalized TNS},
		xlabel near ticks,
		ylabel near ticks,
		legend style={
			draw=none,
			at={(0.5,1.0)},
			anchor=south,
			legend columns=-1,
		}
		]
		
		\pgfplotstableset{
			create on use/x rel/.style={
				create col/expr={
					\thisrow{0}
				}
			},
			create on use/y rel/.style={
				create col/expr={
					\thisrow{1}
				}
			}
		}

		\addplot +[myblue1, line width=1pt, mark=star, mark options={scale=1.0, fill=red},text mark as node=true ] table [x ={er}, y={cpd}] {tnsecgrl.dat};
		\addplot +[mygreen2, line width=1pt, mark=o, mark options={scale=0.5, fill=myorange},text mark as node=true ] table [x ={er}, y={cpd}] {tnsecgagd.dat};
		\addplot +[myred1, line width=1pt, mark=square, mark options={scale=0.5, fill=mymiddle},text mark as node=true] table [x ={er}, y={cpd}] {tnsecgours.dat};
		\legend{RL-Sizer,AGD,Ours}
		
	\end{axis}

\end{tikzpicture} \label{fig:tnsecg}}	\\
	\caption{Normalized TNS and WNS improvement achieved by RL sizer, AGD and our work across different timing requirements.}
	\label{fig:targetopt}
\end{figure*}
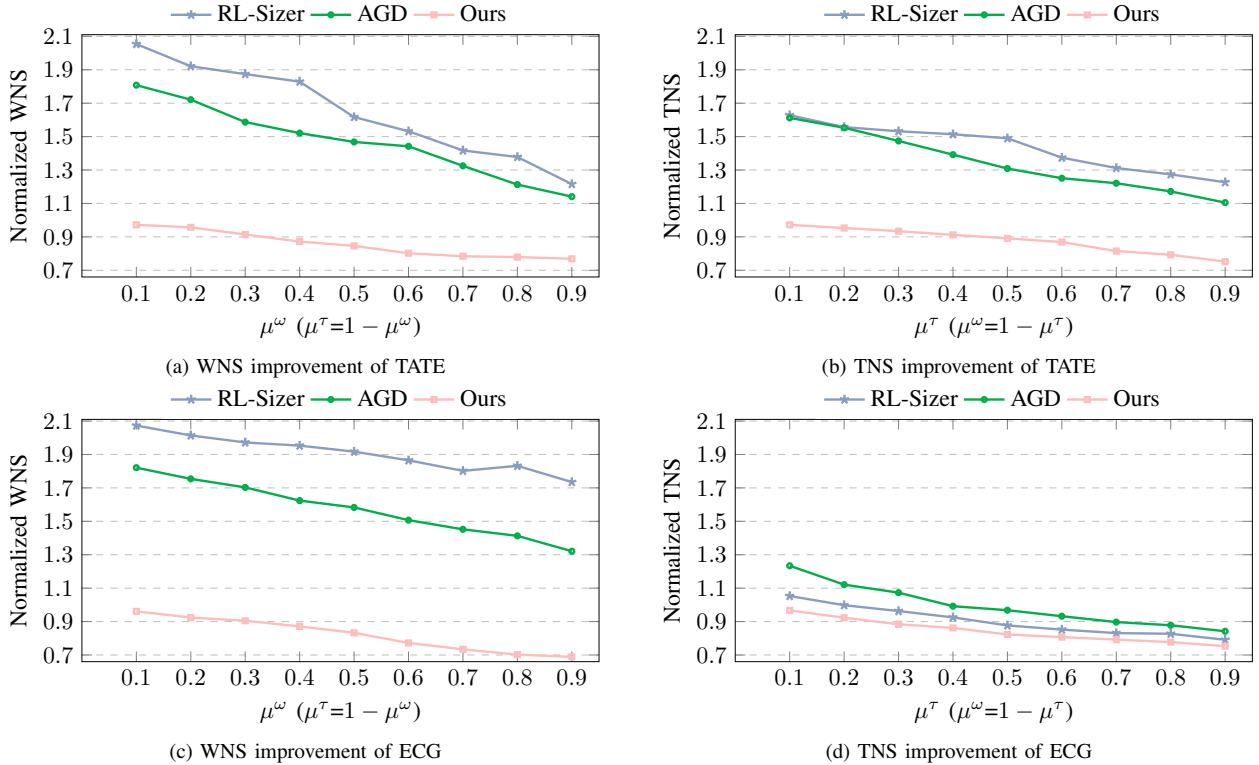
\subsection{Timing Model Accuracy}
The accuracy of our gate-sizing aware timing model to achieve gate-wise TNS and WNS prediction is illustrated in \Cref{tab:model}.
Specifically, the $\text{R}^2$ score (the higher the better) and maximum absolute error $MAE$ (the lower the better) are used to evaluate the performance.
According to the results, they demonstrate that our timing model can accurately predict gate-wise TNS and WNS.
For the training designs, the average $\text{R}^2$ scores and $MAE$ of gate-wise TNS and WNS on all gates are 0.95/6.50$ps$ and 0.97/4.04$ps$.
For the unseen testing designs, the average $\text{R}^2$ scores and $MAE$ of gate-wise TNS and WNS on all gates are 0.94/4.77$ps$ and 0.95/3.14$ps$.
Our proposed framework vastly outperforms all other baseline models on all benchmark circuits for gate-wise TNS and gate-wise WNS prediction.
Compared with DAC22 \cite{guo2022timing} which collects timing information on one single path, timing information on multiple paths is considered jointly in our work.
Compared with DAC23 \cite{wang2023restructure} which collects single-scale physical information, physical information on multi-scale layouts is collected in our model.
The utilized rich information makes our gate sizing-aware timing model more accurate.
The high accuracy helps to improve the timing optimization performance of our work.

\subsection{Timing Performance Improvements}
\Cref{tab:1} demonstrates the timing optimization results of our work and other comparisons after replacement and rerouting.
In summary, our framework achieves an average of 16.29\% and 18.61\% WNS and TNS improvements compared with $ICC2$.
And it also outperforms all other comparisons.
After analyzing the results, we summarize our findings below:
\begin{itemize}
\item Our work can achieve gate sizing to optimize timing on seen and unseen circuits. 
The results suggest that our work can generalize across various designs with different functions and scales without any retraining.
\item We achieve higher timing performance improvements, including TNS, WNS and NVE, with ignorable power consumption costs.
\item Compared with RL-sizer, our gradient-based work can achieve more stable optimization, especially on unseen designs.
\item Compared with Transizer, our work achieves gradient descent optimization. 
$ICC2$ results are used as gradient labels rather than classification labels.
It helps our work outperform $ICC2$ rather than imitate it as Transizer.
\item Compared with AGD, our work achieves better optimization performance benefiting from the multi-modal gate sizing-aware timing model and effective gradient generation and back-propagation.
\end{itemize}

As described in \Cref{sec:target}, our work can optimize timing performance according to different requirements.
It is achieved by adjusting weights for the TNS target $\mu^{\tau}$ and WNS target $\mu^{\omega}$.
\Cref{fig:targetopt} gives results of timing optimization on two Opencore design, including TATE and ECG.
They are achieved by RL-sizer \cite{lu2021rl}, AGD \cite{pham2023agd} and our work when $\mu^{\tau}$ and $\mu^{\omega}$ are set to different values which ranges from 0.1 to 0.9.
According to the results, we summarize some findings below:
\begin{itemize}
	\item Our work outperforms the other two works based on all settings. The results indicate that our work can achieve more stable and efficient optimization across all design spaces.
	\item Larger $\mu^{\tau}$ leads to generating circuits with better TNS optimization, while $\mu^{\omega}$ leads to better WNS optimization.
	It indicates that our work can meet different timing requirements effectively for different applications.
\end{itemize}

\begin{table*}[!t]
	\centering
    \footnotesize
	\caption{Runtime Comparisons.} 
	\label{tab:2}
	{
		\begin{tabular}{|c|c|c|c|c|c|}
			\hline
			\multirow{2}{*}{Circuit} & \multicolumn{5}{c|}{Runtime (min) / Speedup ($\times$)}              \\ \cline{2-6}
			& $ICC2$   & RL-sizer \cite{lu2021rl} & TransSizer \cite{nath2022transsizer} & AGD \cite{pham2023agd}   & Ours  \\ \hline \hline
			DMX                      & 18.72/1.00$\times$   & 33.27/0.56$\times$     & 0.65/28.8$\times$       & 6.23/3.00$\times$   & 3.12/6.01$\times$   \\
			GFX                      & 36.75/1.00$\times$   & 55.13/0.67$\times$     & 0.63/58.0$\times$       & 12.25/3.00$\times$  & 8.17/4.50$\times$   \\
			AC97                     & 7.93/1.00$\times$    & 14.73/0.54$\times$     & 0.52/15.4$\times$       & 2.27/3.50$\times$   & 1.13/7.00$\times$   \\
			VGA                      & 284.97/1.00$\times$  & 313.07/0.91$\times$    & 0.97/294.8$\times$       & 45.15/6.31$\times$  & 32.12/8.87$\times$  \\
			NOVA                     & 153.57/1.00$\times$  & 342.27/0.45$\times$    & 1.63/94.0$\times$       & 41.15/3.73$\times$  & 28.10/5.47$\times$  \\
			TOP                  & 23.45/1.00$\times$   & 45.88/0.51$\times$     & 0.37/64.0$\times$       & 6.12/3.83$\times$   & 4.08/5.74$\times$   \\
			ECG                      & 65.60/1.00$\times$   & 122.12/0.54$\times$    & 1.02/64.5$\times$       & 18.17/3.61$\times$  & 8.07/8.13$\times$   \\
			ETH                      & 22.53/1.00$\times$   & 47.12/0.48$\times$     & 0.83/27.0$\times$       & 7.17/3.14$\times$   & 3.07/7.35$\times$   \\
			USB                      & 17.53/1.00$\times$   & 31.97/0.55$\times$     & 0.58/30.1$\times$       & 4.13/4.24$\times$   & 3.10/5.66$\times$   \\
			TATE                     & 179.37/1.00$\times$  & 267.55/0.67$\times$    & 2.15/83.4$\times$       & 38.08/4.71$\times$  & 27.05/6.63$\times$  \\ \hline \hline
			Ave.                  & 81.04/1.00$\times$   & 127.31/0.59$\times$    & 0.84/76$\times$         & 18.07/3.91$\times$  & 11.80/6.64$\times$ \\ \hline
		\end{tabular}
	}
\end{table*}

\begin{figure*}[t]
	\centering
	\subfloat[]{\centering\includegraphics[height=0.22\linewidth]{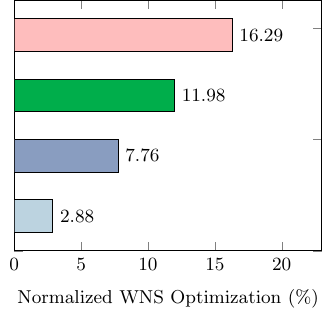} \label{fig:bar1}}
	\subfloat[]{\centering\includegraphics[height=0.22\linewidth]{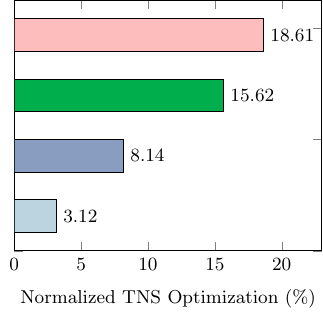} \label{fig:bar2} }
	\subfloat[]{\centering\includegraphics[height=0.22\linewidth]{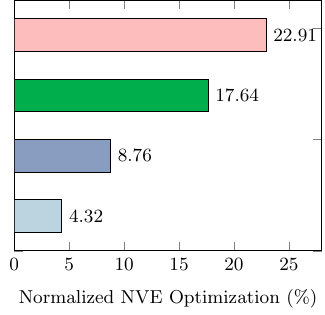} \label{fig:bar3}}
	\subfloat[]{\centering\includegraphics[height=0.22\linewidth]{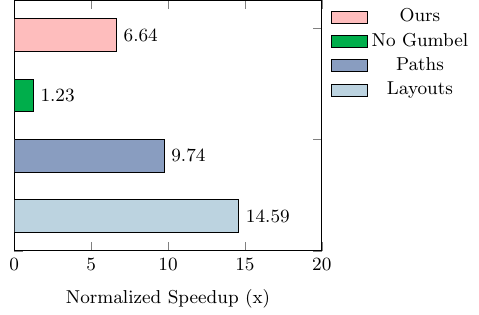} \label{fig:bar4} }
	\caption{{{Comparison among different schemes by (a) WNS optimization, (b) TNS optimization, (c) NVE optimization, and (d) speedup. All these values are normalized by results generated via $ICC2$.}}} 
	\label{fig:bar}
\end{figure*}
\begin{figure}[tb!]
	\centerline{\includegraphics[width=0.4\textwidth]{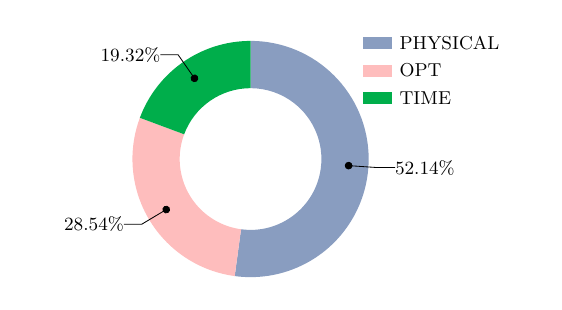}}
	\caption{Runtime breakdown in one gate sizing flow of our work.}
	\label{fig:runtimebreak}
\end{figure}

\subsection{Runtime}
The time-to-market pressure requires the gate sizing work to be effective on large-scale circuits.
The running time of our framework and other comparisons are shown in \Cref{tab:2}.
Compared with timing-consuming $ICC2$ and RL-sizer \cite{lu2021rl}, our work achieves 6.64$\times$ and 11.25$\times$ speedup, respectively.
Compared with TransSizer \cite{nath2022transsizer}, our work achieves much better optimization performance in a reasonable time.
Compared with other gradient descent optimization works AGD \cite{pham2023agd}, our work achieves acceleration benefiting from optimizing critical paths globally and adaptive back-propagation.

In addition, as demonstrated in \Cref{fig:runtimebreak}, the bulk of runtime in one gate sizing flow –about 70\%– is consumed by physical and timing analysis for our work. 
Thus, the overall runtime is predominantly influenced by the convergence speed of the optimization algorithms.
Our framework benefits from an accelerated convergence speed, resulting in faster optimization and more enhanced scalability, which is particularly advantageous for large-scale circuits. 
This acceleration is achieved through adaptive gradient back-propagation and the trained model based on gradient labels.

\subsection{Ablation Study}
In this section, we conduct ablation studies to demonstrate the effectiveness of our proposed work. 
We compare the following schemes:

\noindent (1) \textbf{Layouts-only}: It can utilize multi-scale layout features via physical feature aggregation proposed in \Cref{sec:physical}.

\noindent (2) \textbf{Paths-only}: It can utilize multi-path timing features via timing feature aggregation proposed in \Cref{sec:timing}.

\noindent (3) \textbf{No Gumbel}: It achieves gradient back-propagation without Gumbel-Softmax sampling proposed in \Cref{sec:propa}.

\noindent (4) \textbf{Ours}: It can utilize multi-scale layout features and multi-path timing features.
It achieves gradient back-propagation with Gumbel-Softmax sampling.
This scheme is the final implementation of our work.

As demonstrated in \Cref{fig:bar}, our work are compared with other works by WNS optimization (see \Cref{fig:bar1}), TNS optimization (see \Cref{fig:bar2}), NVE optimization (see \Cref{fig:bar3}) and speedup (see \Cref{fig:bar4}).
According to our results, the most significant improvement is achieved by aggregating multi-path timing features, which is because the timing information on critical paths always is the key to achieve timing optimization.
In addition, the multi-scale physical aggregated features can also help to enhance the optimization performance by capturing the influence from layouts.
As shown in \Cref{fig:bar4}, our final work can achieve 6.64x speedup compared with $ICC2$, which is similar to Paths-only and Layouts-only works.
However, the runtime of the No Gumbel scheme nearly equals the runtime of $ICC2$.
It suggests that adaptive gradient back-propagation through Gumbel-Softmax sampling is efficient to accelerate achieving timing optimization via our work.
In summary, the ablation study validates the benefits of using multi-scale physical features, multi-path timing features, and Gumbel-Softmax sampling.

\section{Conclusion}
This work proposes and implements a learning-driven physically-aware gate sizing framework to achieve timing optimization on large-scale circuits efficiently.
The powerful and efficient optimization is from:
(1) modeling timing optimization and degradation caused by gate-sizing accurately in a multi-modal way via learning timing information on multiple timing paths and physical information on multiple scaled layouts.
(2) generating and back-propagating gradients efficiently to update gate sizes via sizing-oriented straight-through estimator and adaptive sampling.
Experimental results on open-source designs show that our work can achieve 16.29\% and 18.61\% TNS and WNS improvements on average compared with the commercial gate sizing tool.
In addition, it obtains a 6.64$\times$ speedup. 

{
    \bibliographystyle{IEEEtran}
    \bibliography{Top.bib, size.bib}

\begin{thebibliography}{10}
\providecommand{\url}[1]{#1}
\csname url@samestyle\endcsname
\providecommand{\newblock}{\relax}
\providecommand{\bibinfo}[2]{#2}
\providecommand{\BIBentrySTDinterwordspacing}{\spaceskip=0pt\relax}
\providecommand{\BIBentryALTinterwordstretchfactor}{4}
\providecommand{\BIBentryALTinterwordspacing}{\spaceskip=\fontdimen2\font plus
\BIBentryALTinterwordstretchfactor\fontdimen3\font minus \fontdimen4\font\relax}
\providecommand{\BIBforeignlanguage}[2]{{%
\expandafter\ifx\csname l@#1\endcsname\relax
\typeout{** WARNING: IEEEtran.bst: No hyphenation pattern has been}%
\typeout{** loaded for the language `#1'. Using the pattern for}%
\typeout{** the default language instead.}%
\else
\language=\csname l@#1\endcsname
\fi
#2}}
\providecommand{\BIBdecl}{\relax}
\BIBdecl

\bibitem{kahng2011vlsi}
A.~B. Kahng, J.~Lienig, I.~L. Markov, and J.~Hu, \emph{{{VLSI} physical design: from graph partitioning to timing closure}}.\hskip 1em plus 0.5em minus 0.4em\relax Springer, 2011, vol. 312.

\bibitem{kahng2013high}
A.~B. Kahng, S.~Kang, H.~Lee, I.~L. Markov, and P.~Thapar, ``High-performance gate sizing with a signoff timer,'' in \emph{2013 IEEE/ACM International Conference on Computer-Aided Design (ICCAD)}.\hskip 1em plus 0.5em minus 0.4em\relax IEEE, 2013, pp. 450--457.

\bibitem{icc2}
Synopsys, ``{IC Compiler II User Guide},'' \url{ https://www.synopsys.com/implementation-and-signoff/physical-implementation/ic-compiler.html}, 2023.

\bibitem{nath2022transsizer}
S.~Nath, G.~Pradipta, C.~Hu, T.~Yang, B.~Khailany, and H.~Ren, ``{TransSizer}: A novel transformer-based fast gate sizer,'' in \emph{IEEE/ACM International Conference on Computer-Aided Design (ICCAD)}, 2022, pp. 1--9.

\bibitem{livramento2014hybrid}
V.~S. Livramento, C.~Guth, J.~L. Guentzel, and M.~O. Johann, ``A hybrid technique for discrete gate sizing based on lagrangian relaxation,'' \emph{ACM Transactions on Design Automation of Electronic Systems (TODAES)}, vol.~19, no.~4, pp. 1--25, 2014.

\bibitem{sharma2015fast}
A.~Sharma, D.~Chinnery, S.~Bhardwaj, and C.~Chu, ``Fast lagrangian relaxation based gate sizing using multi-threading,'' in \emph{2015 IEEE/ACM International Conference on Computer-Aided Design (ICCAD)}.\hskip 1em plus 0.5em minus 0.4em\relax IEEE, 2015, pp. 426--433.

\bibitem{sharma2019fast}
A.~Sharma, D.~Chinnery, T.~Reimann, S.~Bhardwaj, and C.~Chu, ``Fast {Lagrangian} relaxation-based multithreaded gate sizing using simple timing calibrations,'' \emph{IEEE Transactions on Computer-Aided Design of Integrated Circuits and Systems (TCAD)}, vol.~39, no.~7, pp. 1456--1469, 2019.

\bibitem{roy2015osfa}
S.~Roy, D.~Liu, J.~Um, and D.~Z. Pan, ``{OSFA}: A new paradigm of gate-sizing for power/performance optimizations under multiple operating conditions,'' in \emph{ACM/IEEE Design Automation Conference (DAC)}, 2015, pp. 1--6.

\bibitem{mangiras2022task}
D.~Mangiras, D.~Chinnery, and G.~Dimitrakopoulos, ``Task-based parallel programming for gate sizing,'' \emph{IEEE Transactions on Computer-Aided Design of Integrated Circuits and Systems}, vol.~42, no.~4, pp. 1309--1322, 2022.

\bibitem{daboul2018provably}
S.~Daboul, N.~H{\"a}hnle, S.~Held, and U.~Schorr, ``Provably fast and near-optimum gate sizing,'' \emph{IEEE transactions on computer-aided design of integrated circuits and systems}, vol.~37, no.~12, pp. 3163--3176, 2018.

\bibitem{lu2021rl}
Y.-C. Lu, S.~Nath, V.~Khandelwal, and S.~K. Lim, ``{RL}-sizer: Vlsi gate sizing for timing optimization using deep reinforcement learning,'' in \emph{ACM/IEEE Design Automation Conference (DAC)}, 2021, pp. 733--738.

\bibitem{cheng2023dagsizer}
C.-K. Cheng, C.~Holtz, A.~B. Kahng, B.~Lin, and U.~Mallappa, ``Dagsizer: A directed graph convolutional network approach to discrete gate sizing of vlsi graphs,'' \emph{ACM Transactions on Design Automation of Electronic Systems}, vol.~28, no.~4, pp. 1--31, 2023.

\bibitem{liu2021global}
S.~Liu, Q.~Sun, P.~Liao, Y.~Lin, and B.~Yu, ``Global placement with deep learning-enabled explicit routability optimization,'' in \emph{IEEE/ACM Proceedings Design, Automation and Test in Eurpoe (DATE)}, 2021, pp. 1821--1824.

\bibitem{liu2023concurrent}
S.~Liu, Z.~Wang, F.~Liu, Y.~Lin, B.~Yu, and M.~Wong, ``Concurrent sign-off timing optimization via deep steiner points refinement,'' in \emph{ACM/IEEE Design Automation Conference (DAC)}.\hskip 1em plus 0.5em minus 0.4em\relax IEEE, 2023, pp. 1--6.

\bibitem{pham2023agd}
P.~Pham and J.~Chung, ``{AGD}: A learning-based optimization framework for eda and its application to gate sizing,'' in \emph{ACM/IEEE Design Automation Conference (DAC)}.\hskip 1em plus 0.5em minus 0.4em\relax IEEE, 2023, pp. 1--6.

\bibitem{chen2024ultra}
G.~Chen, Z.~Wang, B.~Yu, D.~Z. Pan, and M.~D. Wong, ``Ultra-fast source mask optimization via conditional discrete diffusion,'' \emph{IEEE Transactions on Computer-Aided Design of Integrated Circuits and Systems}, 2024.

\bibitem{zhu2023l2o}
B.~Zhu, S.~Zheng, Z.~Yu, G.~Chen, Y.~Ma, F.~Yang, B.~Yu, and M.~D. Wong, ``L2o-ilt: Learning to optimize inverse lithography techniques,'' \emph{IEEE Transactions on Computer-Aided Design of Integrated Circuits and Systems}, 2023.

\bibitem{guo2022differentiable}
Z.~Guo and Y.~Lin, ``Differentiable-timing-driven global placement,'' in \emph{Proceedings of the 59th ACM/IEEE Design Automation Conference}, 2022, pp. 1315--1320.

\bibitem{yin2019understanding}
P.~Yin, J.~Lyu, S.~Zhang, S.~Osher, Y.~Qi, and J.~Xin, ``Understanding straight-through estimator in training activation quantized neural nets,'' in \emph{International Conference on Learning Representations (ICLR)}, 2019.

\bibitem{le2022adaste}
H.~Le, R.~K. H{\o}ier, C.-T. Lin, and C.~Zach, ``Adaste: An adaptive straight-through estimator to train binary neural networks,'' in \emph{Proceedings of the IEEE/CVF Conference on Computer Vision and Pattern Recognition}, 2022, pp. 460--469.

\bibitem{yang2022injecting}
Z.~Yang, J.~Lee, and C.~Park, ``Injecting logical constraints into neural networks via straight-through estimators,'' in \emph{International Conference on Machine Learning}.\hskip 1em plus 0.5em minus 0.4em\relax PMLR, 2022, pp. 25\,096--25\,122.

\bibitem{starrc}
Synopsys, ``{StarRC User Guide},'' \url{ https://www.synopsys.com/implementation-and-signoff/signoff/starrc.html}, 2023.

\bibitem{wang2023restructure}
Z.~Wang, S.~Liu, Y.~Pu, S.~Chen, T.-Y. Ho, and B.~Yu, ``{Restructure-Tolerant} timing prediction via multimodal fusion,'' in \emph{ACM/IEEE Design Automation Conference (DAC)}.\hskip 1em plus 0.5em minus 0.4em\relax IEEE, 2023, pp. 1--6.

\bibitem{primetime}
Synopsys, ``{PrimeTime User Guide},'' \url{ https://www.synopsys.com/cgi-bin/imp/pdfdla/pdfr1.cgi?file=primetime-wp.pdf}, 2023.

\bibitem{he2016deep}
K.~He, X.~Zhang, S.~Ren, and J.~Sun, ``Deep residual learning for image recognition,'' in \emph{IEEE Conference on Computer Vision and Pattern Recognition (CVPR)}, 2016, pp. 770--778.

\bibitem{azad2020attention}
R.~Azad, M.~Asadi-Aghbolaghi, M.~Fathy, and S.~Escalera, ``Attention deeplabv3+: Multi-level context attention mechanism for skin lesion segmentation,'' in \emph{European Conference on Computer Vision (ECCV)}.\hskip 1em plus 0.5em minus 0.4em\relax Springer, 2020, pp. 251--266.

\bibitem{kingma2014adam}
D.~P. Kingma and J.~Ba, ``Adam: A method for stochastic optimization,'' \emph{arXiv preprint arXiv:1412.6980}, 2014.

\bibitem{jang2016categorical}
E.~Jang, S.~Gu, and B.~Poole, ``Categorical reparameterization with gumbel-softmax,'' in \emph{International Conference on Learning Representations (ICLR)}, 2016.

\bibitem{guo2022timing}
Z.~Guo, M.~Liu, J.~Gu, S.~Zhang, D.~Z. Pan, and Y.~Lin, ``A timing engine inspired graph neural network model for pre-routing slack prediction,'' in \emph{ACM/IEEE Design Automation Conference (DAC)}, 2022, pp. 1207--1212.

\bibitem{opencore}
OpenCores, \url{ http:///opencores.org/}, 2023.

\end{thebibliography}
}

\end{document}